\documentclass{article}
\usepackage[utf8]{inputenc}
\usepackage{graphicx}
\usepackage{amsmath}
\usepackage[authoryear]{natbib} 
\usepackage{etoolbox}
\usepackage[toc,title]{appendix}
\usepackage{authblk}
\usepackage{multirow}  
\usepackage[most]{tcolorbox}
\usepackage{soul}            
\usepackage{xcolor}
\usepackage{tikz}
\usetikzlibrary{arrows.meta}       
\sethlcolor{yellow!30}       
\usepackage{adjustbox}

\newcommand{\metaphorhl}[1]{{\sethlcolor{purple!20}\hl{#1}}}
\newcommand{\visualhl}[1]{{\sethlcolor{green!20}\hl{#1}}}
\newcommand{\sloganhl}[1]{{\sethlcolor{orange!25}\hl{#1}}}
\usepackage{tikz}
\usetikzlibrary{arrows.meta,calc,positioning,tikzmark} 
\makeatletter
\patchcmd{\NAT@citex}
  {\@citea\NAT@hyper@{\NAT@nmfmt{\NAT@fullname}~\NAT@date}}
  {\@citea\NAT@nmfmt{\NAT@fullname}~\NAT@hyper@{\NAT@date}}
  {}{}

\makeatother
\usepackage{caption}     
\usepackage{hyperref}
\usepackage{amsfonts}
\usepackage{float}
\usepackage{tabularx} 
\usepackage{booktabs} 
\usepackage{csquotes}
\usepackage[none]{hyphenat}
\usepackage{placeins} 
\usepackage{longtable}  
\usepackage{pdflscape}
\usepackage{adjustbox} 
\usepackage{fancyhdr}
\usepackage{colortbl}    
\usepackage{siunitx} 
\usepackage{authblk}
\usepackage[margin=1in]{geometry} 
\usepackage{subcaption}
\usepackage{chngcntr}
\definecolor{lightred}{HTML}{FDE0E0}
\definecolor{lightgreen}{HTML}{E0FDE0}
\usepackage[most]{tcolorbox}      
\tcbuselibrary{listings,breakable} 
\usepackage{xcolor}
\usepackage[T1]{fontenc}
\usepackage[utf8]{inputenc} 
\usepackage{lmodern}       
\tcbset{
  colback=white,      
  colframe=black,     
  colbacktitle=black, 
  coltitle=white,     
  boxrule=0.8pt,      
  arc=3pt,            
  fonttitle=\bfseries 
}

\begin{document}

\title{Galton’s Law of Mediocrity: Why Large Language Models Regress to the Mean and Fail at Creativity in Advertising}

\author[1]{Matt Keon\textsuperscript{*}\thanks{\texttt{mattk@55mv.co}}}
\author[1]{Aabid Karim\textsuperscript{*}\thanks{\texttt{abed.karim@55mv.co}}}
\author[1]{Bhoomika Lohana\textsuperscript{*}\thanks{\texttt{bhoomikalohana70@gmail.com}}}
\author[1]{Abdul Karim\textsuperscript{*}\thanks{\texttt{abdul.karim@55mv.co}}}
\author[2]{Thai Nguyen\thanks{\texttt{minhthai1995@gmail.com}}}
\author[3]{Tara Hamilton\thanks{\texttt{t.hamilton@westernsydney.edu.au}}}
\author[1]{Ali Abbas\thanks{\texttt{aliabbas.2810@gmail.com}}}

\affil[1]{55mv Research Lab}
\affil[2]{Monash University}
\affil[3]{School of Engineering, Western Sydney University.}

\maketitle

\renewcommand{\thefootnote}{\fnsymbol{footnote}}
\footnotetext[1]{These authors contributed equally.}

\begin{abstract}

Large language models (LLMs) generate fluent text yet often default to safe, generic phrasing, raising doubts about their ability to handle creativity. We formalize this tendency as a Galton-style regression to the mean in language and evaluate it using a creativity stress test in advertising concepts.  When ad ideas were simplified step by step, creative features such as metaphors, emotions, and visual cues disappeared early, while factual content remained, showing that models favor high-probability information. When asked to regenerate from simplified inputs, models produced longer outputs with lexical variety but failed to recover the depth and distinctiveness of the originals. We combined quantitative comparisons with qualitative analysis, which revealed that the regenerated texts often appeared novel but lacked true originality. Providing ad-specific cues such as metaphors, emotional hooks and visual markers improved alignment and stylistic balance, though outputs still relied on familiar tropes. Taken together, the findings show that without targeted guidance, LLMs drift towards mediocrity in creative tasks; structured signals can partially counter this tendency and point towards pathways for developing creativity-sensitive models.
\end{abstract}

\section{Introduction}

Over the past decade, LLMs have undergone rapid and transformative development. Ground-breaking innovations such as Word2Vec \citep{mikolov2013efficient}, Transformers \citep{vaswani2017attention}, and scaling laws \citep{brown2020language} have reshaped natural language processing (NLP), problem solving, question answering, and computer vision tasks \citep{li2025visual}, \citep{gunter2024apple}, \citep{team2023gemini}. 
However, these models consistently struggle in one critical area: genuine creativity \citep{ismayilzada2024evaluating}, because they are trained to maximise next-token likelihood and thus privilege high-probability continuations and suppress rare or unconventional ones \citep{lu2024benchmarking}.  This bias explains what \citep{wenger2025we} call creative homogeneity: even diverse models converge on strikingly similar responses in divergent thinking tasks, a pattern they warn could narrow rather than expand human idea space. Similarly, \citep{anderson2024homogenization} report that while LLMs generate grammatically polished narratives, they offer little variation in perspective or style, which constrains expression when used as co-creators. Moreover, the assumption that larger datasets inherently yield deeper insights has not held true; as training corpora expand, this statistical bias becomes more pronounced \citep{chen2024from}, \citep{liu2025not}, \citep{muennighoff2023scaling}.
\vspace{0.5 em}

We call this tendency Galton’s Law of Mediocrity. First introduced by \citep{galton1886regression}, the law describes the statistical principle of regression towards the mean. When unusually extreme traits appear in a population, for example, exceptional height or intelligence, they are unlikely to persist across generations and instead drift back towards the population average. We adapt this principle to LLMs: because they are trained to maximise next-token likelihood, they privilege high-frequency, statistically probable patterns while suppressing rare or unconventional ones. Consequently, even as models scale with larger datasets, their outputs converge on safe and predictable phrases, leaving little room for genuinely novel or creative expression. \citep{pasquinelli2021nooscope} argue that this tendency reflects a deeper limitation: models do not invent but remix dominant cultural patterns, which can give the illusion of creativity without genuine novelty. Empirical studies reinforce this point. \citep{haase2025has} find that the same model can swing between highly novel and bland responses depending on the prompt, while \citep{kumar2025human} report that LLMs generate many ideas quickly in terms of fluency but fall short on originality and real-world usefulness. Together, these findings suggest that scale brings polish, but not the kind of originality that sustains long-run creativity. Alignment efforts add another layer to this regression. \citep{mohammadi2024creativity} shows that reinforcement learning from human feedback (RLHF), while improving safety and politeness, lowers entropy and reduces surprising continuations, revealing a trade-off between reliability and generative richness. \citep{franceschelli2024creativity} further argues that much of what appears as creativity in LLMs reflects surface fluency rather than true conceptual innovation.   
\vspace{0.5 em}

From Ada Lovelace’s claim that machines “originate nothing” to Turing’s suggestion that novelty might emerge through recombination, the question of whether machines can truly generate new ideas has remained central for nearly two centuries. This long-standing debate becomes especially vivid in advertising, a domain where emotional resonance, cultural relevance, and originality are everything. Unlike abstract tests of narrative or brainstorming, advertising provides concrete examples of how a single metaphor, image, or slogan can define success, making it an ideal setting to probe whether LLMs truly generate creativity or regress towards mediocrity. Classic campaigns like Nike’s “Just Do It” \citep{embryo_justdoit}, Apple’s “Think Different” \citep{forbes_thinkdifferent}, and Volkswagen’s “Lemon” \citep{michael2025vwlemon} are remembered not because they recombined stock phrases, but because they blended novel narratives and striking images in ways that spoke to the time. When we ask today’s LLMs to create an eco‑friendly slogan, however, they often reply with something generic like “Go Green!” instead of inventing culturally nuanced metaphors. Even when models produce surprising combinations, such as an ad that mixes coffee and astronomy, the result often lacks value without human interpretation. This reflects what \citep{tian2024large}  identify in narrative generation, where LLMs struggle with emotional depth, and what \citep{gu2024interesting} observe in scientific brainstorming, where models contribute only moderately novel ideas unless curated by humans. Turning a raw output into a culturally resonant campaign like Starbucks’ Galaxy Blend therefore still requires a human creative director.  To illustrate this gap more concretely, we focus on two famous car ads as our motivation, that stood out: Volkswagen’s “Darth Vader” commercial \citep{sanburn2015time_vwforce} and Volvo Trucks’ “Epic Split” \citep{contagious2014volvo_live_tests}. The former pairs a family sedan’s remote‑start feature with a child in a Darth Vader costume, creating a humorous twist; the latter uses an bold stunt to show a truck’s stability. These advertisements demonstrate how much success depends on a single visual joke or stunt, an aspect that current language models often fail to include. These two advertisements, along with still images are provided in the  Appendix  \ref{appendix:a} for reference. That’s why we chose advertising as our evaluation setting: it exposes the gap between the lively, surprising content humans create and the cautious outputs from today’s LLMs.
\vspace{0.5 em}

Our goal is to understand, in a systematic way, why current LLMs regress towards the average in creative domains such as advertising. We pose three questions:
\begin{enumerate}
    \item When we gradually simplify ads, which part lasts longest, the generic product details or the unusual creative highlights?
    \item If we give the model only the bare facts, can it fill in the creative flourishes or does it fall back on clichés?
    \item Can explicit creative cues (“markers”) improve the recovery of originality during regeneration?
    
\end{enumerate}

 To answer them, we design a structured, two-phase empirical study to be conducted using our 55mV idea database (1000+ advertising ideas). The database covers a wide range of advertising concepts that show key aspects of creativity, such as telling a story, creating an emotional connection, sparking visual imagination, and using short, memorable slogans. We run a staged forgetting experiment, asking a state‑of‑the‑art LLM to simplify each ad in several steps and measuring how quickly creative content disappears. Second, we invert the process: starting from the compressed outputs of phase 1, we prompt the same model to expand them back into full ads and analyse the kinds of details it invents or omits. 
\vspace{0.5 em}

Across all SOTA models, we find that forgetting erases creative content first; expansions produce surface-level novelty but fail to restore originality; and marker-driven prompting (prompts include creative markers/concepts) improves diversity by reintroducing visual, emotional, and slogan-like cues, though models still recycle familiar tropes. Our contributions are following:
\begin{itemize}
    \item We formalise Galton’s Law of Mediocrity for LLMs and identify the factors that enforce it.
    \item We propose a two-phase creativity stress test, combining controlled forgetting and expansion, and introduce a hybrid evaluation framework that integrates quantitative metrics (cosine similarity, METEOR, entropy, n-gram uniqueness) with qualitative human + LLM judgment. The details of these metrics are provided in Section \ref{sec:evaluation_metrics}.
   
\end{itemize}

The rest of this paper is organised as follows: Section 2 describes our experimental design, Section 3 presents the results, and Section 4 discusses implications and future directions.

\section{Methodology}
\label{sec:methodology}

In advertising, creativity is not just about filling space with words or visuals; it is about producing ideas that stand out, surprise, and connect emotionally with audiences. Yet, large language models often default to safe, predictable outputs, demonstrating creative homogeneity that produces ideas that appear creative but lack the diversity and novelty found in human creativity \citep{wenger2025we}. To study this limitation, our experiments begin with the 55mV Idea Database, which contains more than 1000 original advertising concepts selected for their creative richness. These concepts included metaphorical phrasing, emotional tone, visual imagination, and cultural specificity, characteristics that typically challenge generative models. Representative examples from the dataset are presented in the  Appendix~\ref{appendix:c1}. All model generations are produced using a consistent sampling configuration (fixed temperature and sampling settings) to ensure comparability. We evaluate five state-of-the-art LLM families: GPT-4o, GPT-5 (main and thinking variants), GPT-OSS-120B, Gemini-2.5-Pro, and Claude Sonnet 4. The experiment is structured in two phases: Phase 1 (forgetting), where ads are contracted through staged reductions, and Phase 2 (expansion), where models attempt to regenerate creative detail from compressed inputs.

 \subsection{Phase 1 Forgetting:} In this phase, we prompt LLMs to compress original advertising ideas through three levels of reduction: mild, moderate, and extreme. At each level, the model rewrites the ad with fewer details, always starting from the original ad concept. As shown in Appendix ~\ref{appendix:b}, Figure ~\ref{fig:contraction-prompt}, the prompts do not specify which details should be removed or retained. This open-ended design lets us observe what the model naturally chooses to keep and what it lets go of, providing insight into its own forgetting tendencies.

\begin{itemize}
    \item \textbf{Mild reduction ($\approx$35\%):} most of the ad’s structure and creative features remain, with some details removed.
    \item \textbf{Moderate reduction ($\approx$70\%):} only the central idea is preserved, while most creative layers are removed.
    \item \textbf{Extreme reduction ($\approx$95\%):} the ad is compressed into one or two sentences containing only the bare core message.
\end{itemize}

This process is repeated across three independent runs per model to capture both consistent forgetting patterns and natural variation. The outcome is three versions of each original ad (three reductions), which form the dataset for our later quantitative and qualitative analyses. The goal of this phase is to test whether, as ads are simplified, creative elements such as metaphors, humor, and emotional cues are preserved or disappear faster than the factual product message.
   
 \vspace{0.5em}
 
Figure \ref{fig:phase1_forgetting_example} in appendix \ref{appendix:c} shows how the forgetting process operates on an original advertisement about car sensors. The original version combines multiple creative elements: a metaphor (“tree people”), \textbf{humour/surprise} (sudden movement), and \textbf{sensory detail} (kids, cyclists, pedestrians). These layers create a vivid and imaginative message. As the text is progressively reduced, these features disappear in a predictable order. At the \textbf{mild stage (35\%)}, the metaphor remains but with fewer details. At the \textbf{moderate stage (70\%)}, the metaphor is weakened, and sensory detail is largely absent. At the \textbf{extreme stage (95\%)}, only the core product message survives (“Car sensors detect what you can’t see”), with creativity fully collapsed.

\vspace{0.5em}

Table~\ref{tab:phase1_features} is based on a close reading of the reduced outputs in figure \ref{fig:phase1_forgetting_example}, noting how different creative parts survived at each level.  Metaphors, humour, sensory details, and story structure were judged by whether they still shaped the ad or had faded out, while the product message was treated as a basic factual check. This highlights not only what remains but also the order in which creativity tends to disappear.

\begin{table}[H]
\centering
\begin{tabular}{|l|c|c|c|c|}
\hline
Feature & Original & Mild (35\%) & Moderate (70\%) & Extreme (95\%) \\
\hline
Metaphor (tree people) & Strong & Present & Reduced & Absent \\
Humour / Surprise & Present & Present & Weak & Absent \\
Sensory detail (scaring, hiding) & Rich & Reduced & Minimal & Absent \\
Narrative structure & Complex & Simplified & Flattened & Collapsed \\
Core product message (car sensors) & Present & Present & Present & Present (only) \\
\hline
\end{tabular}
\caption{Creative feature tracking across abstraction levels.}
\label{tab:phase1_features}
\end{table}

\subsection{Phase 2 Expansion:} 
Phase 2 reverses the forgetting procedure by testing whether lost creativity can be rebuilt through step-by-step expansion. This process follows a chained expansion design. Starting from the extreme forgetting versions produced in Phase 1, each advertisement is expanded sequentially: Mild → Moderate → Extreme. In this setup, the Phase 1 extreme output serves as the base concept for the mild expansion; the mild version is then used as input to produce the moderate expansion; and the moderate version in turn feeds into the extreme expansion. At each stage, the newly generated text becomes the input for the next, creating a chained process rather than restarting from the original. This design allows us to observe whether creativity re-emerges gradually as details are layered back in.

\vspace{0.5em}
The same set of models used in Phase 1 are evaluated under identical prompting conditions, and the same four metrics are applied: cosine similarity for semantic fidelity, METEOR for surface overlap, entropy for lexical diversity, and 4-gram uniqueness for structural novelty. In every case, the original ads serve as a fixed benchmark, providing a reference point for recovery.

\vspace{0.5em}
\begin{itemize}
 \item\textbf{Plain Expansion:} The model receives the extreme forgotten version of an ad as input and expands it across mild, moderate, and extreme stages without additional guidance. This establishes a baseline for how much creativity can be recovered autonomously. An illustrative example of this process is provided in Appendix \ref{appendix:c3}, and the corresponding prompt is shown in Appendix \ref{appendix:b}.
 
\vspace{0.5em}

  \item\textbf{Marker-Driven Expansion:} Alongside the extreme forgotten version, the model is given 3–4 creative markers extracted from the original ad. These markers are short, ad-specific phrases (e.g., emotional cues, metaphors, visual details, brand references, or slogans) extracted by an LLM with a structured prompt. Sample list of  markers is provided in Appendix \ref{appendix:c5}. Unlike a generic list, each set of markers is unique to its advertisement. They are introduced progressively: one or two at mild expansion, most at moderate, and all at extreme. This setup tests whether explicit creative cues help the model restore lost originality rather than defaulting to generic filler. An illustrative example of this process is included in Appendix \ref{appendix:c4}, and the corresponding prompt is shown in Appendix \ref{appendix:b}.
   \end{itemize}
\vspace{0.5em}

\subsection{Evaluation Metrics}
\label{sec:evaluation_metrics}
Evaluating creativity requires multiple perspectives, since no single metric captures its full scope. We consider four complementary measures, cosine similarity, METEOR, entropy, and 4-gram uniqueness, chosen to reflect semantic fidelity, surface overlap, lexical diversity, and structural novelty. Together, these measures provide a multidimensional view of how creativity degrades during forgetting and re-emerges during expansion. For each phase, metric values are first computed for every individual advertisement at each level of forgetting or expansion (mild, moderate, extreme). These values are then averaged across the 1,045 ads in the dataset. Finally, we report the mean ± standard deviation across three independent runs per model, with the original ad serving as the baseline for comparison. The detailed computation of these metrics is provided in Appendix \ref{appendix:d}
\vspace{0.5em}

\begin{enumerate}
    \item \textbf{Cosine Similarity:} We compute cosine similarity between embedding representations of the generated text and the original advertisement to assess semantic fidelity. This indicates whether the core idea is preserved or diverges. In forgetting, lower similarity signals semantic drift, while higher values indicate retention. In expansion, increases suggest recovery of the original concept, whereas decreases imply elaboration without alignment.
    \item \textbf{METEOR}: METEOR measures surface-level correspondence, accounting for synonyms and word order flexibility \citep{banerjee-lavie-2005-meteor}. Unlike BLEU, which penalizes variation, METEOR tolerates creative rephrasings while still capturing recognizable overlap with the original. We use it to assess whether compressed or expanded texts maintain a recognizable connection to the original creative concept, even when expressed in different wording.
    \item \textbf{Entropy} We compute token-level entropy over the generated text to quantify lexical diversity. Higher entropy reflects richer and more varied word use, while lower entropy reflects reduced variety and repetitive phrasing. We use entropy because diversity in word choice is a key marker of creative expression.
    \item \textbf{4-gram Uniqueness} This metric measures the proportion of novel 4-grams in the generated text relative to the original, capturing structural novelty at the lexical level. We use it to assess whether the model re-expresses ideas with fresh phrasing instead of relying on repeated or conservative wording. We use 4-grams because they are long enough to capture meaningful patterns of phrasing, but not so long that they become too rare to compare reliably.
\end{enumerate}

\section{Experiments and Results}
\subsection{Phase 1 Forgetting: Quantitative Analysis}

To account for randomness in sampling, we ran each experiment three times per model. Variation across runs was consistently minimal: standard deviations remained below 0.03 for cosine similarity, METEOR, entropy, and n-gram uniqueness at all levels. This indicates that the observed trends are highly stable across models. For clarity of presentation, we report the results from the last run (Run 3) in the main text. The full mean ± SD values across runs are provided in Appendix  \ref{appendix:d} Table \ref{tab:phase1_stability}.

\begin{figure}[H]
    \centering
    \scriptsize
    \includegraphics[width=\linewidth]{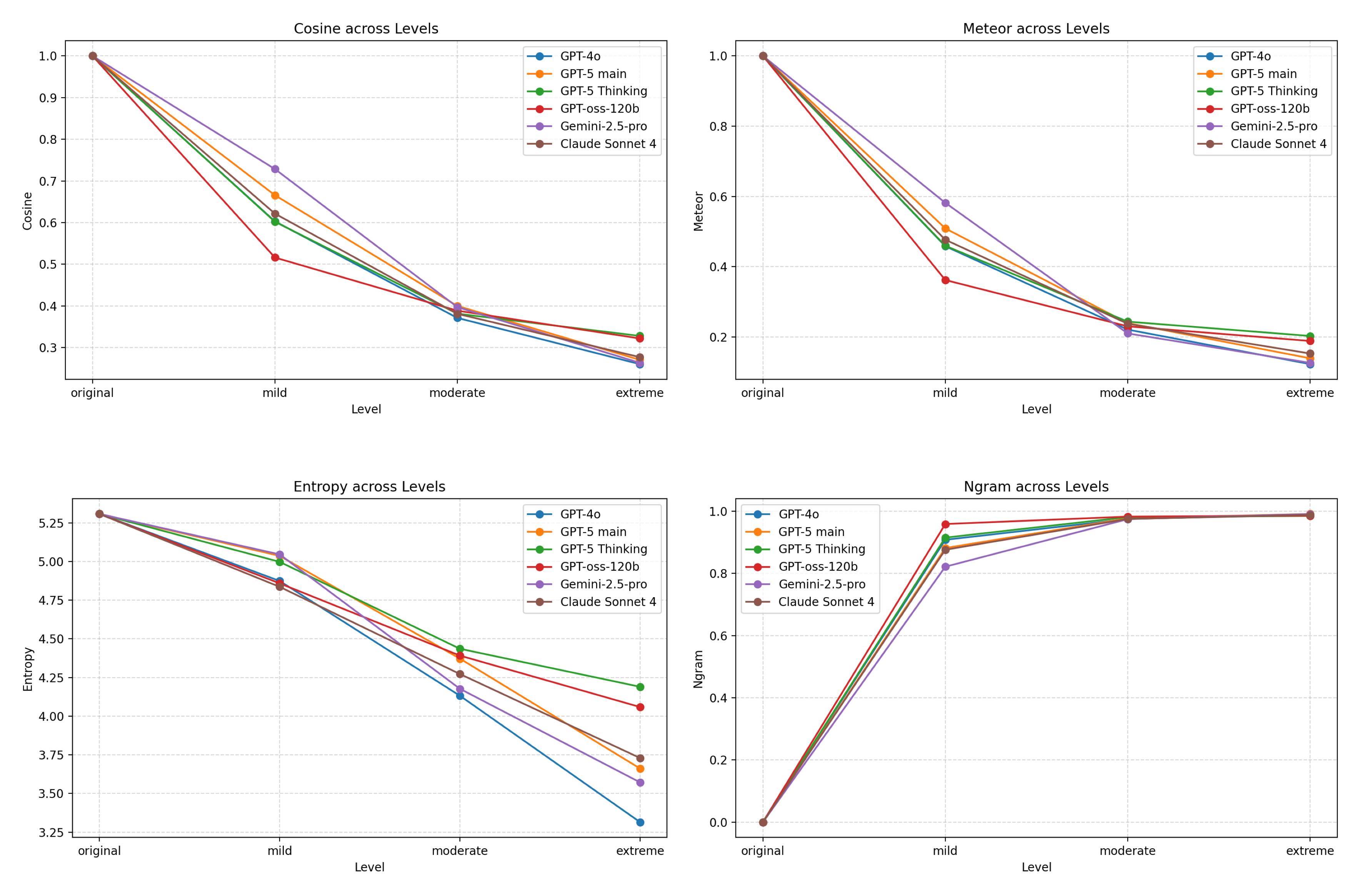}
    \caption{Phase 1 Forgetting Trends Across Models}
    \label{fig:phase1_four_metrics}
\end{figure}

\vspace{1em}
For each model and forgetting level, we first computed cosine similarity, METEOR, entropy, and 4-gram uniqueness for every ad relative to its original version. The values reported in  Table \ref{tab:metrics-all} and Figure \ref{fig:phase1_four_metrics}, are the averages across all 1,045 ads, giving a single aggregate score per metric. 

\vspace{1em}
As shown in Table \ref{tab:metrics-all} and Figure \ref{fig:phase1_four_metrics}, both cosine similarity and METEOR fell sharply as forgetting increased, averaged across all ad concepts. This confirmed that the generated outputs moved further away from the originals, both in meaning and in recognizable surface wording. At the mild stage, models differ in how quickly they diverge: Gemini-2.5-pro remains closest to the original (cosine 0.73, METEOR 0.58),  

\begin{table}[H]
\centering
\scriptsize
\renewcommand{\arraystretch}{1.1}
\begin{tabular}{|l|l|c|c|c|c|}
\hline
\textbf{Model} & \textbf{Level} & \textbf{Cosine} & \textbf{METEOR} & \textbf{Entropy} & \textbf{N-gram} \\
\hline
\multirow{4}{*}{GPT-4o}
 & Original & 1.00 & 1.00 & 5.31 & 0.00 \\
 & Mild     & 0.60 & 0.46 & 4.87 & 0.91 \\
 & Moderate & 0.37 & 0.22 & 4.13 & 0.97 \\
 & Extreme  & 0.26 & 0.12 & 3.31 & 0.99 \\
\hline
\multirow{4}{*}{GPT-5 main}
 & Original & 1.00 & 1.00 & 5.31 & 0.00 \\
 & Mild     & 0.67 & 0.51 & 5.04 & 0.88 \\
 & Moderate & 0.40 & 0.24 & 4.37 & 0.98 \\
 & Extreme  & 0.27 & 0.14 & 3.66 & 0.99 \\
\hline
\multirow{4}{*}{GPT-5 Thinking}
 & Original & 1.00 & 1.00 & 5.31 & 0.00 \\
 & Mild     & 0.60 & 0.46 & 5.00 & 0.91 \\
 & Moderate & 0.38 & 0.24 & 4.45 & 0.98 \\
 & Extreme  & 0.33 & 0.20 & 4.19 & 0.98 \\
\hline
\multirow{4}{*}{GPT-oss-120b}
 & Original & 1.00 & 1.00 & 5.31 & 0.00 \\
 & Mild     & 0.52 & 0.36 & 4.86 & 0.96 \\
 & Moderate & 0.39 & 0.23 & 4.39 & 0.98 \\
 & Extreme  & 0.32 & 0.19 & 4.06 & 0.99 \\
\hline
\multirow{4}{*}{Gemini-2.5-pro}
 & Original & 1.00 & 1.00 & 5.31 & 0.00 \\
 & Mild     & 0.73 & 0.58 & 5.05 & 0.82 \\
 & Moderate & 0.40 & 0.21 & 4.18 & 0.97 \\
 & Extreme  & 0.26 & 0.13 & 3.57 & 0.99 \\
\hline
\multirow{4}{*}{Claude Sonnet 4}
 & Original & 1.00 & 1.00 & 5.31 & 0.00 \\
 & Mild     & 0.62 & 0.48 & 4.84 & 0.88 \\
 & Moderate & 0.38 & 0.24 & 4.27 & 0.97 \\
 & Extreme  & 0.28 & 0.15 & 3.73 & 0.99 \\
\hline
\end{tabular}
\caption{Phase 1 averaged results across models and forgetting levels.}
\label{tab:metrics-all}
\end{table}

while GPT-oss-120b drifts away the fastest (cosine 0.52, METEOR 0.36). GPT-4o and the GPT-5 variants fall between these extremes, preserving some connection but already losing large parts of the original. By the moderate stage, these differences vanish and every model converges near the same low averages (cosine = 0.37–0.40, METEOR = 0.21–0.24). At this stage, most of the distinctive creative framing has already been stripped away, leaving only simplified restatements of the central idea. Because the outputs are reduced to this generic backbone, model-specific strengths no longer create visible differences, and the systems converge towards the same type of summary. At the extreme stage, small gaps reopen: GPT-5 Thinking retains slightly more fidelity (cosine 0.33, METEOR 0.20), while Gemini collapses furthest (cosine 0.26, METEOR 0.13). The overall pattern showed that forgetting followed a collapse pattern: models begin with different trajectories, but once creative detail is stripped away, they converge on similar generic outputs.

\vspace{1em}
Entropy and 4-gram uniqueness reveal the same dynamic in reverse. Entropy declines steadily from ~5.3 in the originals to ~3.3–4.2 at the extreme, showing how vocabulary variety eroded as detail was stripped away. GPT-5 Thinking and GPT-oss-120b hold on to more lexical diversity, while GPT-4 collapses most severely, producing the most repetitive phrasing. In contrast, 4-gram uniqueness rises sharply, climbing from near zero to ~0.99 for all models at the extreme. This indicated that while the compressed outputs looked new on the surface, they were only superficially novel, paraphrased shells stripped of metaphor, humor, and sensory richness. The mild stage again shows differences: GPT-oss-120b departs earliest, introducing new phrasing almost immediately, while Gemini stays closer to the original. But by the moderate and extreme levels, every model saturates near the ceiling, suggesting a universal drift towards reworded but flattened text.

\vspace{1em}

Taken together, these quantitative patterns provide clear evidence for creativity regression through forgetting. The models begin differently, some shorten conservatively, others paraphrase aggressively, but as forgetting increases, they all converge on the same endpoint: semantically thinned, lexically repetitive, and stylistically generic outputs. Instead of compressing in ways that sharpen or distill creative ideas, forgetting consistently drives the models towards mediocrity. Still, at the extreme stage, subtle differences remain: GPT-5 Thinking and GPT-OSS-120B retain slightly higher similarity and lexical diversity than the other models. This raised a key question: did these traces of richness provide an advantage when models are asked to expand ideas again? This naturally led to the question of whether better retention during forgetting enabled stronger expansions.

\subsubsection{Phase1: Qualitative Analysis}

\begin{figure}[H]
\centering
\scriptsize

\begin{tikzpicture}[remember picture,>=Latex]

\node[inner sep=0pt] (adbox) {
  \begin{tcolorbox}[
    enhanced, sharp corners,
    colback=white, colframe=black,
    colbacktitle=black, coltitle=white,
    title=Original Advertisement,
    width=0.9\linewidth, boxsep=4pt
]
  Social media campaign for a new line of vegetarian meals for a restaurant or dinner range. 
  There are many misconceptions about what vegetarians eat. In this campaign we are 
  \metaphorhl{flipping the script on traditional vegetarian dishes}. 
  The posts are styled like a fashion shoot. 
  \visualhl{The models are the vegetarian dishes}. 
  They are styled with clothes made from the ingredients in the dish. For example, PASTA is wearing a dress made 
  from different types of pasta, topped with cheese. Another dish is wearing a hat made of the ingredients in the dish. 
  The models are striking poses and looking fierce. With a tagline below, 
  \sloganhl{\emph{Vegetarian, but make it fashion.}}
  \end{tcolorbox}
};

\node[below=0.5cm of adbox] (markersbox) {
  \begin{tcolorbox}[
    enhanced, sharp corners,
    colback=white, colframe=black,
    colbacktitle=black, coltitle=white,
    title=Extracted Markers (with categories),
    width=0.9\linewidth, boxsep=4pt
]
  \begin{itemize}\itemsep2pt
    \item \textbf{Metaphorical}: flipping the script on traditional vegetarian dishes
    \item \textbf{Visual}: the models are the vegetarian dishes
    \item \textbf{Slogan-like}: ``Vegetarian, but make it fashion.''
  \end{itemize}
  \end{tcolorbox}
};

\draw[very thick,-{Latex[length=3mm]}] (adbox.south) -- (markersbox.north);

\end{tikzpicture}

\caption{Marker-Driven: Original advertisement with extracted markers (bottom).}
\label{fig:ad_markers_arrow}
\end{figure}
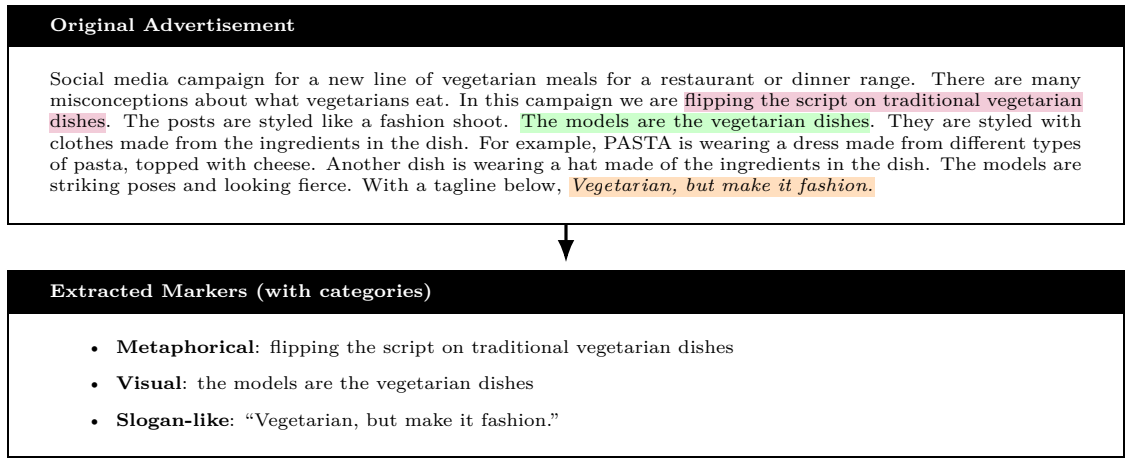

To complement the quantitative metrics, we conducted a qualitative marker-based analysis of 25 advertisements sampled from the dataset. For each ad, we extracted 3–4 short “creative markers”, phrases capturing distinctive features such as metaphor, emotional tone, visual detail, brand-specific cues, or slogan-like phrasing taken as a creative hook that conveys an ad’s originality as shown in  Figure~\ref{fig:ad_markers_arrow}. To carry this out, we used a constrained prompt that identifies 3–4 markers per ad, each limited to ten words or fewer to maintain precision and comparability. We then tracked the survival of these markers across mild, moderate, and extreme forgetting, using a hybrid evaluation: an LLM provides initial judgments (present, partly present, absent), which were manually verified for accuracy. The sample chosen ensured consistency while maintaining scalability. This hybrid approach drew on established frameworks such as MT-Bench \citep{zheng2023judgingllmasajudgemtbenchchatbot} and recent work on LLMs as qualitative judges \citep{chiang2023largelanguagemodelsalternative}, adapted here for creative advertising content.
\vspace{1em}

Figure \ref{fig:ad_markers_arrow} illustrates how this process works with one example: a social campaign reframing vegetarian dishes as high-fashion. From the original text, markers such as “flipping the script on traditional vegetarian dishes” (metaphor), “models are the vegetarian dishes” (visual), and “vegetarian, but make it fashion” (slogan) are extracted as the key creative elements. These markers provide a consistent reference point for assessing what survives as detail is stripped away.

\vspace{1em}

\begin{figure}[H]
    \centering
    \scriptsize
    \begin{subfigure}[t]{0.49\textwidth}
        \centering
        \includegraphics[width=\linewidth]{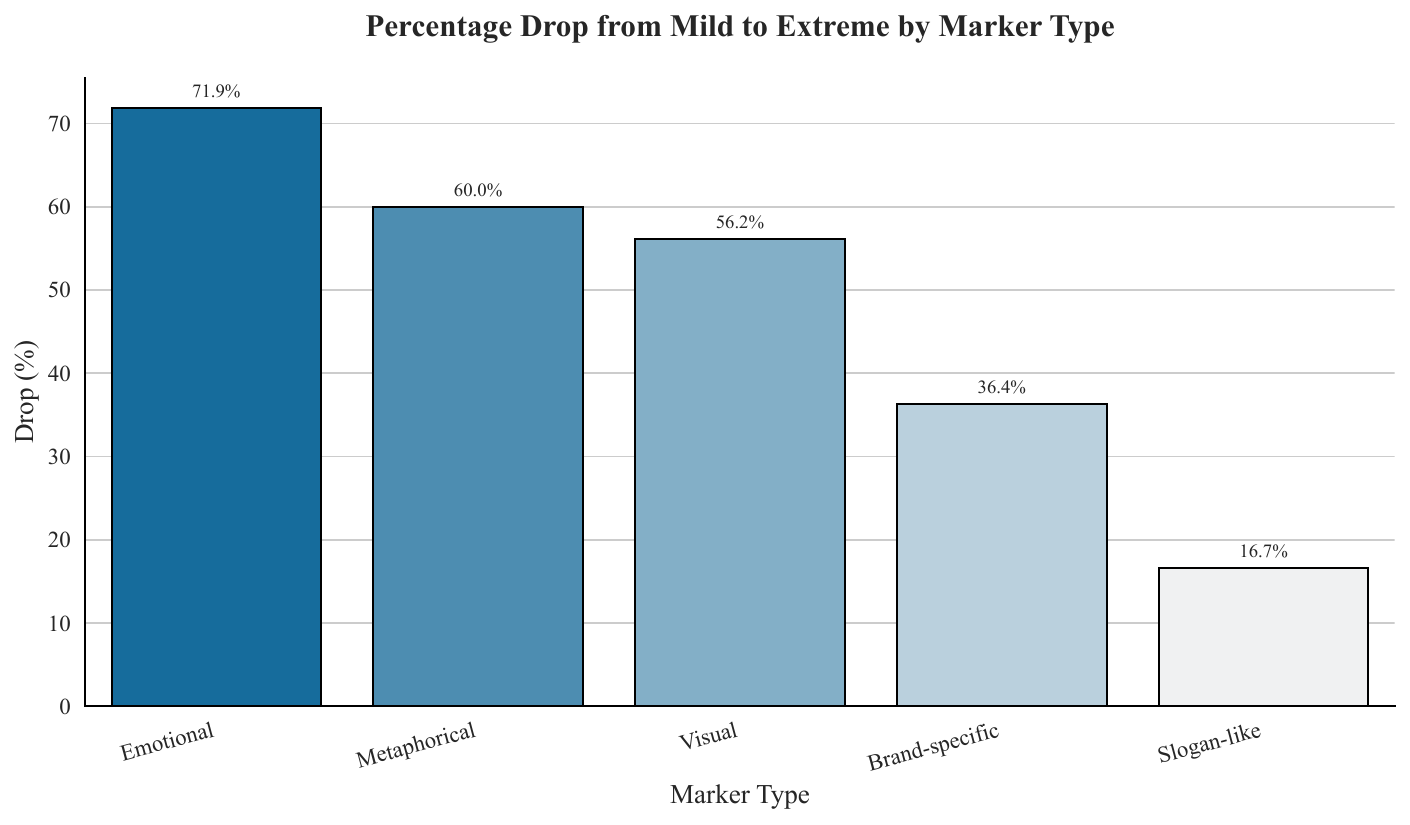}
        \caption{Percentage drop in marker presence from mild to extreme forgetting levels.}
        \label{fig:marker-drop}
    \end{subfigure}\hfill
    \begin{subfigure}[t]{0.49\textwidth}
        \centering
        \includegraphics[width=\linewidth]{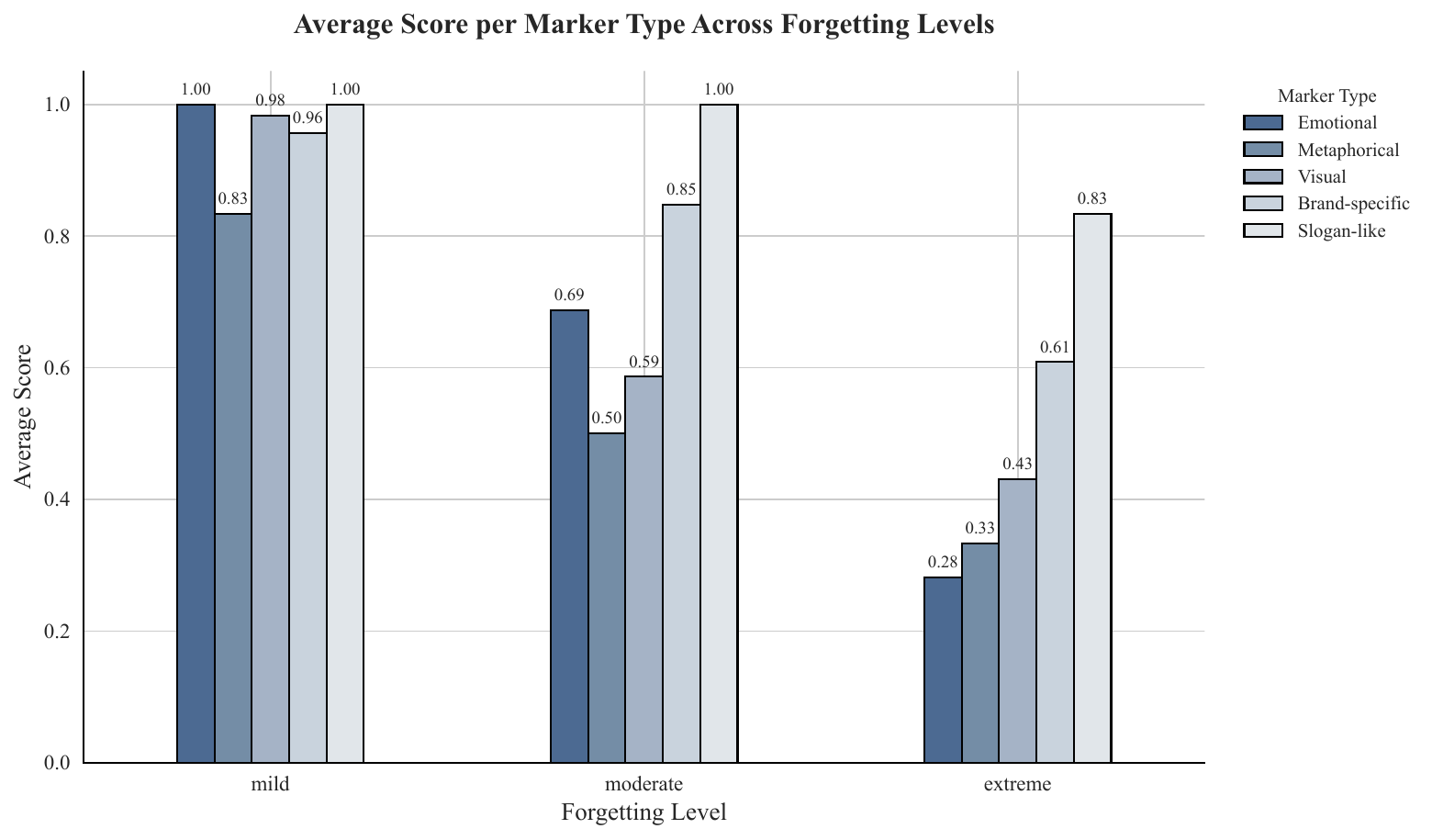}
        \caption{Average score per marker type at each forgetting level.}
        \label{fig:marker-scores}
    \end{subfigure}
    \caption{Comparison of marker performance.}
    \label{fig:marker-comparison}
\end{figure}

The results reveal a structured pattern of loss. Emotional and metaphorical markers are the most fragile, declining by more than 60–70\% between mild and extreme forgetting as shown in the Figure ~\ref{fig:marker-drop}. Visual details also erode sharply (~56\%), while brand-specific cues and slogans remain more resilient, dropping by only ~30–40\%. 

At the extreme stage, the average survival scores for emotional and metaphorical devices fall to 0.28 and 0.33 respectively, compared to 0.61 for brand cues and 0.83 for slogan as shown in the Figure ~\ref{fig:marker-scores}. In practice, this means that the expressive hooks that make ads memorable are the first to vanish, leaving only stripped-down taglines and functional product statements.
\vspace{1em}

Together, these qualitative findings reinforce the quantitative patterns. Forgetting does not selectively distill creativity into sharper form but instead erases its most imaginative components, emotion, metaphor, and vivid imagery, while leaving behind only the safe, formulaic anchors of brand identity. This structured collapse consolidates our hypothesis that LLMs outputs under compression regress towards mediocrity.

\subsection{Phase 2: Expansion}
\label{sec:phase2}
\subsubsection{Plain Expansion: Quantitative Analysis}

In Phase 2 we test whether the creative detail lost during forgetting can be rebuilt. Expansions began from the extreme forgotten versions produced in Phase 1, which served as the starting point for regeneration. The baseline for evaluation was always the original advertisement, and at each expansion level the regenerated text was compared back to this original. The same four metrics used in Phase 1 are applied here: for each advertisement we calculated values at every level, then averaged these across all 1,045 ads. In the main text we report the results from the final run, while Appendix~\ref{appendix:d2} Table \ref{tab:phase2_plain_stability} presents the mean ± SD across all three runs. Standard deviations remained below 0.02 for nearly all metrics, confirming that the observed patterns were stable and not an artifact of sampling variance.

\begin{figure}[H]
  \centering
  \scriptsize
  \includegraphics[width=\linewidth]{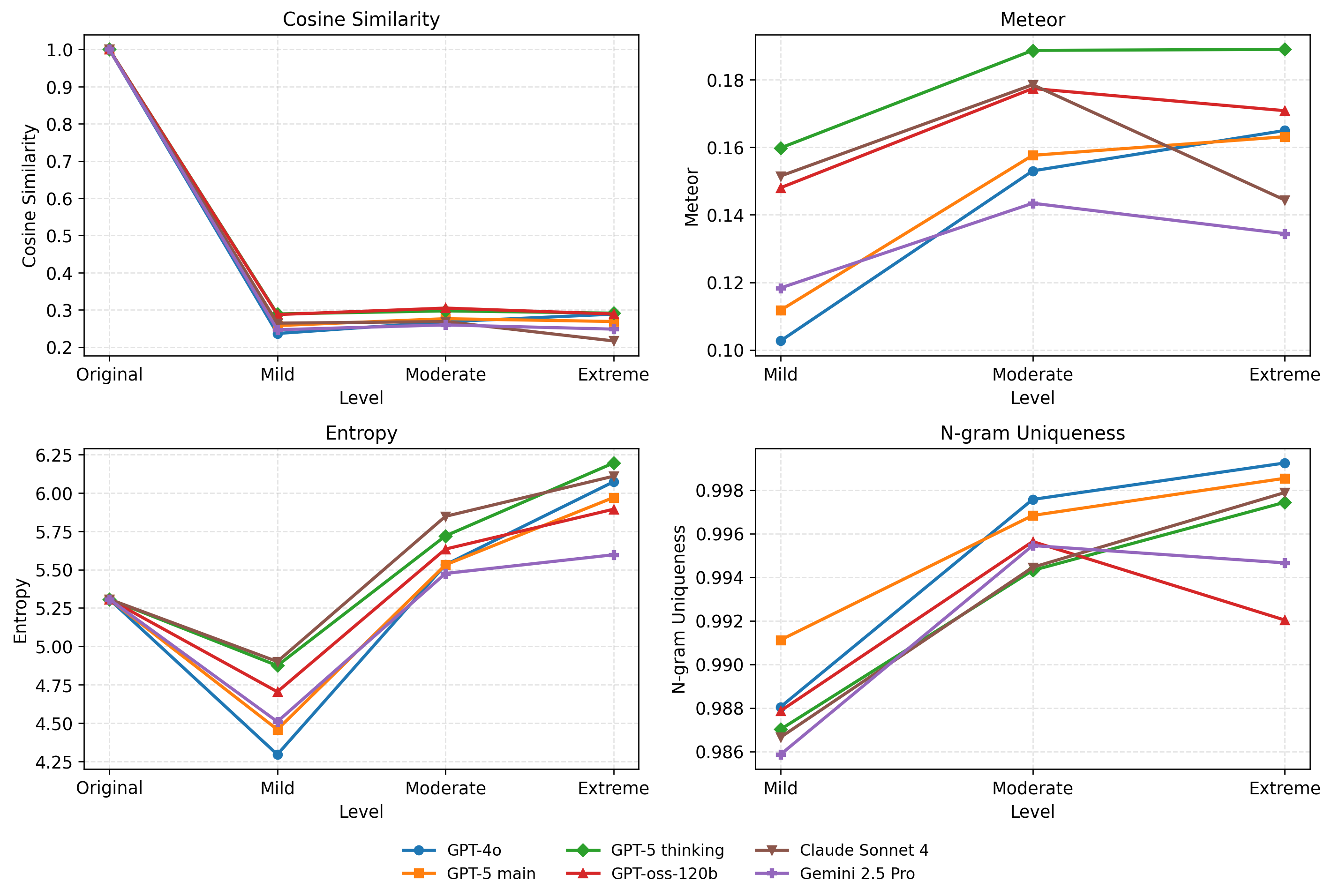}
  \caption{Phase-2 trends across Model.}
  \label{fig:phase2_2x2_mixed}
\end{figure}

\vspace{1em}
As shown in the Table~\ref{tab:phase2_results_levels} and Figure~\ref{fig:phase2_2x2_mixed}, cosine similarity and METEOR improved a little from mild to moderate expansion, but remained low across all models at the extreme level, with scores between 0.22 (Claude Sonnet-4) and 0.29 (GPT-4o, GPT-5 Thinking, GPT-oss-120b). METEOR values follow the same pattern, plateauing at 0.13–0.19 depending on the model, again well below the benchmark of 1.0. Together, these two metrics indicate that expansions introduce detail but recover only a fraction of the original meaning and surface overlap.

\vspace{1em}
Entropy, by contrast, exceeded the original baseline, rising from 5.31 in the ads to between 5.60 (Gemini-2.5 Pro) and 6.20 (GPT-5 Thinking) at the extreme level. This consistent increase across models signaled richer lexical variety in the expanded outputs than in the source ads. Similarly, n-gram uniqueness reached saturation, with nearly all models producing values of 0.99–1.00 at the extreme stage, meaning that the regenerated ads are composed almost entirely of novel phrasing. At first glance this looked like strong originality, but when read together with the low cosine similarity and METEOR, it showed that the novelty is mostly in surface wording. The expansions sound fresh, yet they do not recover much of the original meaning and creative content.

\vspace{1em}

Taken together, these results highlighted a consistent quantitative profile: expansions generated outputs that are lexically varied and almost entirely novel in form, but their semantic and surface alignment with the originals remained limited. Within this overall pattern, differences between models were still visible. As shown in Figure \ref{fig:phase2_2x2_mixed}, GPT-5 Thinking and GPT-OSS-120B performed better, higher on similarity and overlap, while also producing more lexically diverse expansions. GPT-4o and GPT-5 Main follow at a mid-range level, whereas Claude and Gemini consistently trail behind. Appendix \ref{appendix:e} includes a cross-model example where expansions of the same advertisement are compared, illustrating how these quantitative differences appeared qualitatively. These outcomes reinforced the Phase 1 observation: models that retained slightly more information during forgetting were also better positioned to expand the original creative ideas.

\begin{table}[H]
\centering
\scriptsize
\setlength{\tabcolsep}{6pt}
\renewcommand{\arraystretch}{1}
\begin{tabular}{|l|l|c|c|c|c|}
\hline
\textbf{Model} & \textbf{Level} & \textbf{Cosine} & \textbf{METEOR} & \textbf{Entropy} & \textbf{N-gram} \\
\hline
\multirow{4}{*}{GPT-4o}
  & Original & 1.00 & 1.00 & 5.31 & 0.00 \\
  & Mild     & 0.24 & 0.10 & 4.30 & 0.99 \\
  & Moderate & 0.27 & 0.15 & 5.53 & 1.00 \\
  & Extreme  & 0.29 & 0.16 & 6.07 & 1.00 \\
\hline
\multirow{4}{*}{GPT-5 main}
  & Original & 1.00 & 1.00 & 5.31 & 0.00 \\
  & Mild     & 0.26 & 0.11 & 4.46 & 0.99 \\
  & Moderate & 0.28 & 0.16 & 5.53 & 1.00 \\
  & Extreme  & 0.27 & 0.16 & 5.97 & 1.00 \\
\hline
\multirow{4}{*}{GPT-5 Thinking}
  & Original & 1.00 & 1.00 & 5.31 & 0.00 \\
  & Mild     & 0.29 & 0.16 & 4.87 & 0.99 \\
  & Moderate & 0.30 & 0.19 & 5.72 & 0.99 \\
  & Extreme  & 0.29 & 0.19 & 6.20 & 1.00 \\
\hline
\multirow{4}{*}{GPT-oss-120b}
  & Original & 1.00 & 1.00 & 5.31 & 0.00 \\
  & Mild     & 0.29 & 0.15 & 4.70 & 0.99 \\
  & Moderate & 0.30 & 0.18 & 5.63 & 1.00 \\
  & Extreme  & 0.29 & 0.17 & 5.89 & 0.99 \\
\hline
\multirow{4}{*}{Gemini 2.5 Pro}
  & Original & 1.00 & 1.00 & 5.31 & 0.00 \\
  & Mild     & 0.25 & 0.12 & 4.51 & 0.99 \\
  & Moderate & 0.26 & 0.14 & 5.48 & 1.00 \\
  & Extreme  & 0.25 & 0.13 & 5.60 & 0.99 \\
\hline
\multirow{4}{*}{Claude Sonnet 4}
  & Original & 1.00 & 1.00 & 5.31 & 0.00 \\
  & Mild     & 0.26 & 0.15 & 4.90 & 0.99 \\
  & Moderate & 0.27 & 0.18 & 5.85 & 0.99 \\
  & Extreme  & 0.22 & 0.14 & 6.11 & 1.00 \\
\hline
\end{tabular}
\caption{Phase-2 averaged results across models and levels.}
\label{tab:phase2_results_levels}
\end{table}

\vspace{2 em}
\subsubsection{Marker driven expansion: Quantitative Analysis}

To quantify recovery under marker-driven expansion, we again compared the regenerated ads at each level against the originals using the same four metrics as in Phase 1. Results were averaged across all 1,045 ads, with each experiment repeated three times per model. Full mean ± SD values are reported in Appendix \ref{appendix:d}, Table \ref{tab:phase2_marker_stability}. Across all metrics, the standard deviations remained below 0.02 in almost all the runs, reducing the chances of randomness.
\vspace{0.5 em}

\begin{figure}[H]
    \centering
    \scriptsize
    \includegraphics[width=\textwidth]{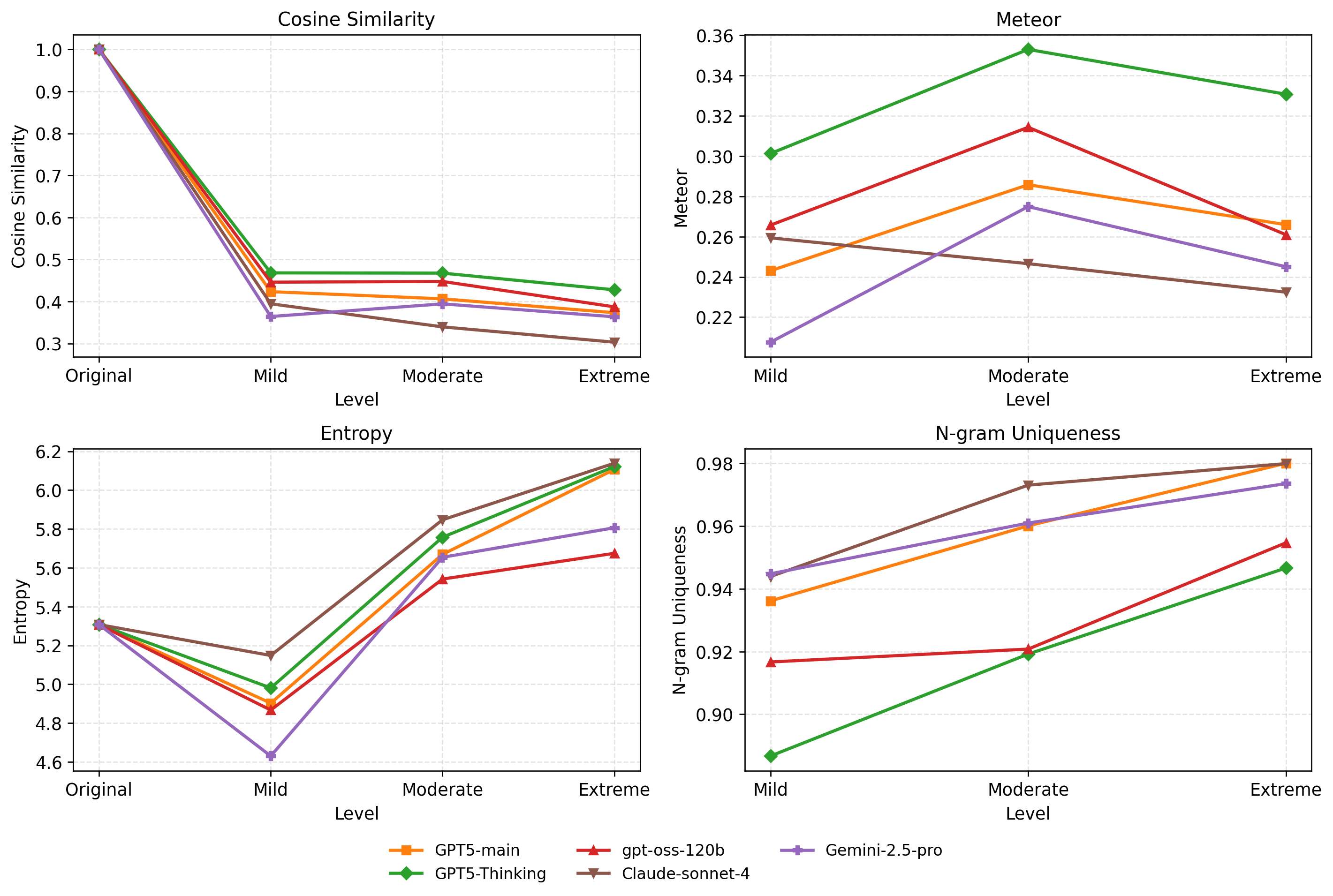}
    \caption{Phase-2 Marker-Driven trends across levels.}
    \label{fig:Marker_driven_metric}
\end{figure}

We again saw in the Table   \ref{tab:phase2_avg_metrics} and Figure ~\ref{fig:Marker_driven_metric}, clear patterns across the four metrics. Cosine similarity scores rose noticeably relative to plain expansion, reaching values between 0.24 (Gemini-2.5 Pro) and 0.37 (GPT-5 main) at the extreme stage. While these values remained well below the original baseline (1.0), the increase over plain expansion suggested that marker cues help models preserve more of the original semantic content. METEOR followed a similar trend, with extreme values ranging from 0.13 (Gemini) to 0.26 (GPT-5 main). The improvement was modest but consistent: marker-driven outputs showed higher surface-level overlap with the originals than those generated without guidance.

\begin{table}[H]
\centering
\scriptsize
\setlength{\tabcolsep}{6pt}
\begin{tabular}{|l|l|c|c|c|c|}
\hline
\textbf{Model} & \textbf{Level} & \textbf{Cosine} & \textbf{METEOR} & \textbf{Entropy} & \textbf{N-gram} \\
\hline
\multirow{4}{*}{GPT-4o}
 & Original & 1.00 & 1.00 & 5.31 & 0.00 \\
 & Mild     & 0.40 & 0.23 & 4.73 & 0.94 \\
 & Moderate & 0.35 & 0.22 & 5.44 & 0.98 \\
 & Extreme  & 0.33 & 0.21 & 5.85 & 0.99 \\
\hline
\multirow{4}{*}{GPT-5 main}
 & Original & 1.00 & 1.00 & 5.31 & 0.00 \\
 & Mild     & 0.42 & 0.24 & 4.91 & 0.94 \\
 & Moderate & 0.41 & 0.28 & 5.66 & 0.96 \\
 & Extreme  & 0.37 & 0.26 & 6.10 & 0.98 \\
\hline
\multirow{4}{*}{GPT-5 Thinking}
 & Original & 1.00 & 1.00 & 5.31 & 0.00 \\
 & Mild     & 0.47 & 0.30 & 4.97 & 0.89 \\
 & Moderate & 0.47 & 0.35 & 5.75 & 0.92 \\
 & Extreme  & 0.43 & 0.33 & 6.11 & 0.95 \\
\hline
\multirow{4}{*}{gpt-oss-120b}
 & Original & 1.00 & 1.00 & 5.31 & 0.00 \\
 & Mild     & 0.44 & 0.26 & 4.90 & 0.92 \\
 & Moderate & 0.46 & 0.32 & 5.61 & 0.93 \\
 & Extreme  & 0.40 & 0.28 & 5.81 & 0.96 \\
\hline
\multirow{4}{*}{Gemini-2.5-pro}
 & Original & 1.00 & 1.00 & 5.31 & 0.00 \\
 & Mild     & 0.36 & 0.20 & 4.63 & 0.95 \\
 & Moderate & 0.39 & 0.28 & 5.64 & 0.96 \\
 & Extreme  & 0.36 & 0.24 & 5.75 & 0.97 \\
\hline
\multirow{4}{*}{Claude Sonnet 4}
 & Original & 1.00 & 1.00 & 5.31 & 0.00 \\
 & Mild     & 0.39 & 0.26 & 5.01 & 0.92 \\
 & Moderate & 0.34 & 0.24 & 5.84 & 0.97 \\
 & Extreme  & 0.30 & 0.23 & 6.13 & 0.98 \\
\hline
\end{tabular}
\caption{Phase-2 marker-driven averaged results across models and levels.}
\label{tab:phase2_avg_metrics}
\end{table}

\vspace{1em}
Entropy values at the extreme level were slightly higher than in plain expansion, ranging from 5.60 (Gemini) to 6.13 (Claude Sonnet 4). This suggested that reintroducing markers not only anchors the outputs semantically but also stimulates greater lexical variety. Importantly, the increases were consistent across all models, indicating that marker cues systematically encouraged richer expression. N-gram uniqueness remained very high (0.95–0.99 at extreme), showing that while models integrate markers, they still rely on fresh phrasings rather than copying from the original ads.

\vspace{1 em}
Model-level differences also emerged. GPT-5 Thinking and GPT-oss-120b consistently achieved higher METEOR (0.28–0.33) and entropy (5.8–6.1) scores, suggesting they used the marker guidance more effectively to balance overlap and diversity. By contrast, Gemini-2.5 Pro and Claude Sonnet 4 showed lower semantic fidelity (cosine 0.24–0.30) despite producing outputs with strong lexical variety, reflecting a tendency to elaborate creatively but drift from the original concept. GPT-5 Thinking, with cosine 0.43 and METEOR 0.33 at the extreme, stood out as the model that most successfully reconnected expansions with the originals under marker guidance, while also producing the lowest n-gram uniqueness compared to other models. This indicated that it not only recaptured meaning but also retained a closer structural resemblance to the original ideas under marker guidance.

\vspace{1em}
Taken together, the quantitative results showed that marker-driven expansion improved recovery compared to plain expansion. The extreme versions regained more semantic alignment and recognizable overlap, while still producing highly novel text. The gains are not uniform across models, but the consistent upward shifts in cosine similarity, METEOR, and entropy suggested that structured cues provided a measurable advantage when expanding forgotten ads.

\subsection{Comparison: Plain vs. Marker-Driven Expansion}

When comparing the two expansion strategies side by side, a consistent advantage emerged for the marker-driven condition. Cosine similarity and METEOR scores were systematically higher under marker guidance, indicating that expansions anchored by creative markers remained more semantically aligned with the originals and preserved more recognizable overlap. Entropy was also higher under marker-driven expansion, reflecting richer lexical variety without simple repetition. At the same time, n-gram uniqueness was lower with markers, showing that while vocabulary broadens, phrasing remained focused around key cues rather than pushing towards maximal surface novelty. These effects were consistent across runs. By contrast, plain expansion drifted towards surface-level novelty, showing higher n-gram uniqueness but lower cosine similarity and METEOR, yielding lexically diverse outputs that are less connected to the original ad ideas. These results suggested that providing models with structured creative cues reduced the burden of unconstrained generation, allowing them to recover both alignment and diversity more effectively.

\subsection{Phase2: Qualitative Analysis}

While the quantitative metrics in Phase 2 established how much the regenerated ads diverged from the originals, they leave open a key question: what kinds of creativity are the models actually producing? Cosine similarity and METEOR scores tell us about semantic alignment, and entropy and n-gram uniqueness measure lexical variety, but none of these capture the nature of the new ideas being introduced in generated ads. To address this gap, we conducted a qualitative analysis of the extreme expansions, the stage where the model stretches furthest from the original text. We aimed to characterize the stylistic forms of creativity that emerge and to see whether these felt genuinely novel or instead reflect familiar advertising tropes.
\vspace{0.1em}

We randomly sampled 100 ads from the full dataset to ensure broad coverage without topical bias. For each case, we compared the original advertisement with its extreme expansion and extracted only emergent ideas, novel elements that appeared in the expansion but are absent from the original. For instance, if an ad originally described "a tea brand that helps you relax after work" and the expansion introduced "a sanctuary where every sip whispers calm into your soul," the metaphorical phrase "sanctuary where every sip whispers calm" is counted as an emergent idea. Each emergent idea was then coded into one of four categories adapted from advertising and linguistics research: Metaphorical, Emotional, Visual, or Slogan-like. This method provided a systematic way to track how LLMs express creativity across ads.
\vspace{0.1em}

\begin{figure}[H]
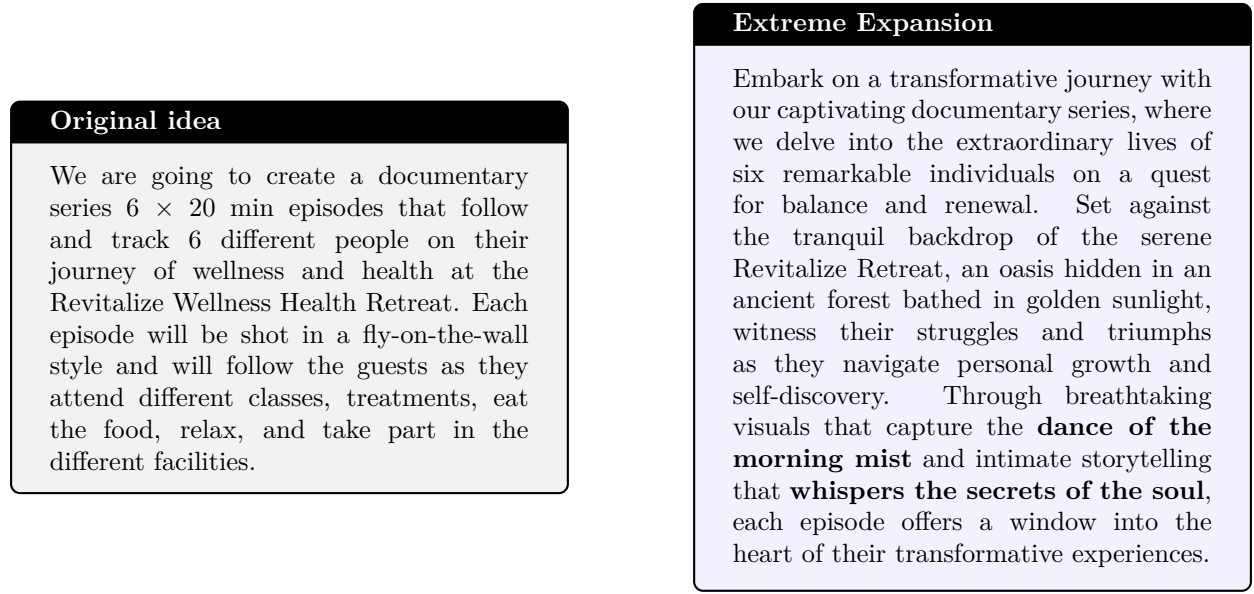

\centering
\begin{minipage}{0.45\textwidth}
\begin{tcolorbox}[colback=gray!10, colframe=black, title=Original idea]
We are going to create a documentary series 6 × 20 min episodes that follow 
and track 6 different people on their journey of wellness and health at the 
Revitalize Wellness Health Retreat. Each episode will be shot in a fly-on-the-wall 
style and will follow the guests as they attend different classes, treatments, 
eat the food, relax, and take part in the different facilities.
\end{tcolorbox}
\end{minipage}
\hfill
\begin{minipage}{0.45\textwidth}
\begin{tcolorbox}[colback=blue!5, colframe=black, title=Extreme Expansion]

Embark on a transformative journey with our captivating documentary series, 
where we delve into the extraordinary lives of six remarkable individuals 
on a quest for balance and renewal. Set against the tranquil backdrop of the 
serene Revitalize Retreat, an oasis hidden in an ancient forest bathed in 
golden sunlight, witness their struggles and triumphs as they navigate 
personal growth and self-discovery. Through breathtaking visuals that 
capture the \textbf{dance of the morning mist} and intimate storytelling 
that \textbf{whispers the secrets of the soul}, each episode offers a 
window into the heart of their transformative experiences.
\end{tcolorbox}
\end{minipage}
\caption{Comparison of an Original Ad and its Extreme Expansion.}
\label{fig:AD_EXAMPLE}
\end{figure}

\begin{figure}[H]
    \centering
    \begin{subfigure}[t]{0.45\textwidth}
        \centering
        \includegraphics[width=\linewidth]{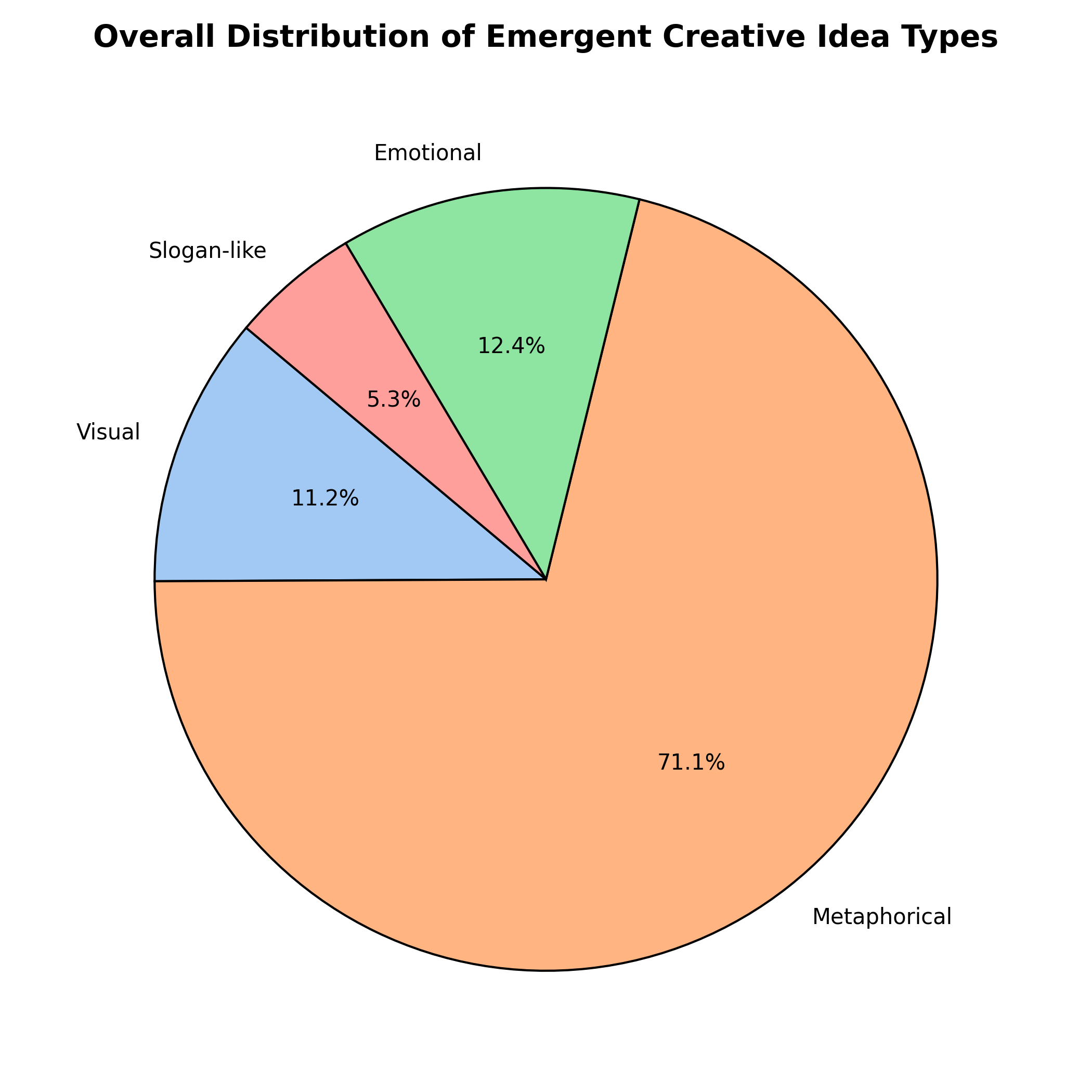}
        \caption{Distribution of emergent idea types.}
        \label{fig:emergent-idea-types}
    \end{subfigure}\hfill
    \begin{subfigure}[t]{0.45\textwidth}
        \centering
        \includegraphics[width=\linewidth]{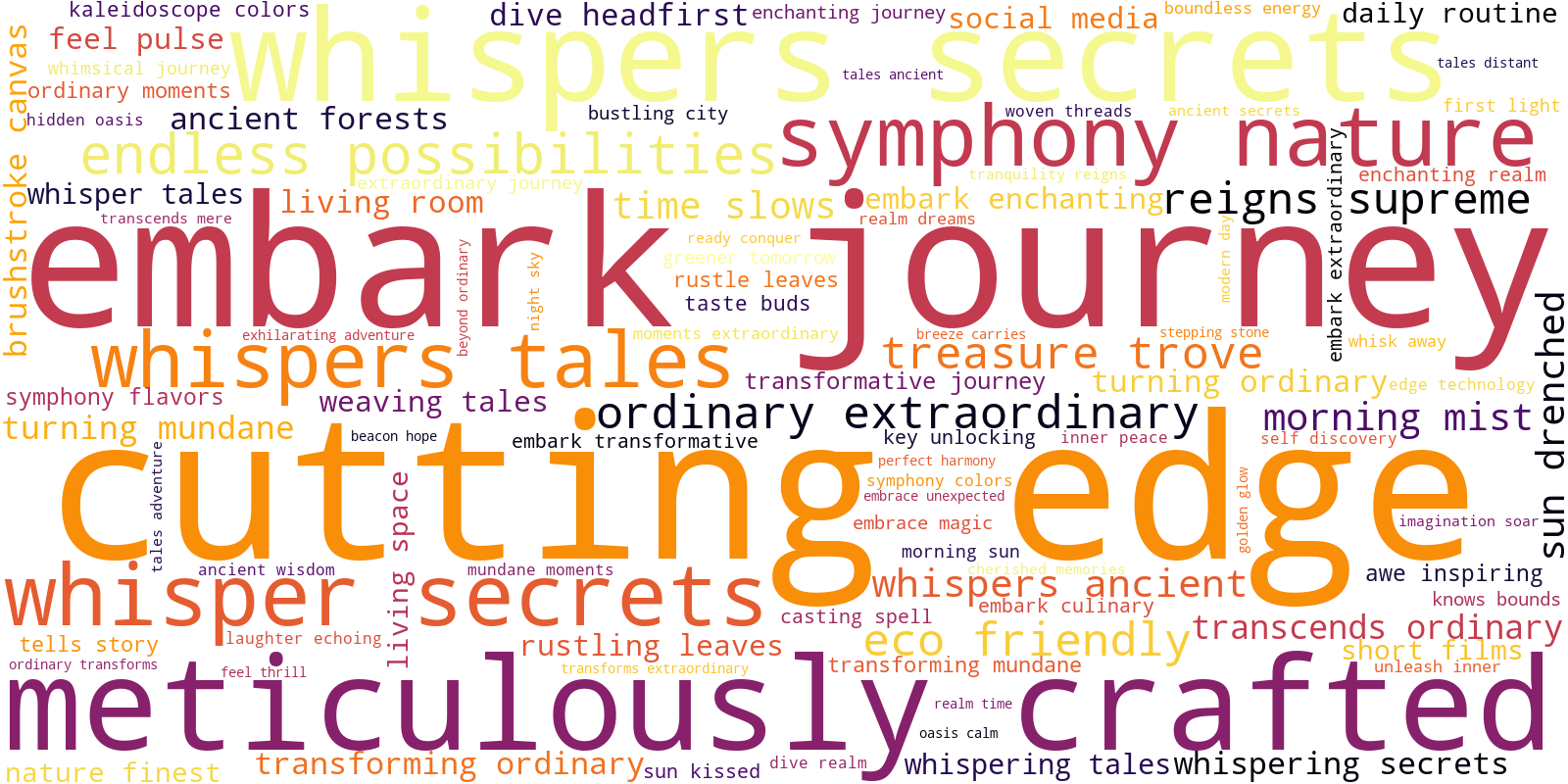}
        \caption{Bigram word cloud of extreme expansions.}
        \label{fig:bigram-wordcloud}
    \end{subfigure}
    \caption{Comparison of emergent idea types and word clouds from extreme expansions.}
    \label{fig:emergent-vs-wordcloud}
\end{figure}

The results showed a clear imbalance. As Figure \ref{fig:emergent-idea-types} illustrates, 71\% of emergent ideas are metaphorical, far outnumbering the other categories. Emotional (12.5\%), Visual (11.3\%), and Slogan-like (5.2\%) ideas are much less frequent. This suggested that when unconstrained, models default to figurative elaboration as their dominant creative mode. Metaphors can help, but the bias points to a small, repetitive style set. Figure \ref{fig:AD_EXAMPLE}  illustrates this dynamic: the original ad idea for a wellness documentary is relatively straightforward, whereas the extreme expansion transforms it into a florid narrative filled with imagery such as "an oasis hidden in an ancient forest" and "the dance of the morning mist that whispers the secrets of the soul." These flourishes were poetic, but they echoed patterns seen repeatedly across the dataset, often leaning on familiar clichés such as nature, mystery, and spirituality rather than introducing genuinely new ideas.
\vspace{0.5em}

\begin{figure}[H]
    \centering
    \includegraphics[width=0.75\textwidth]{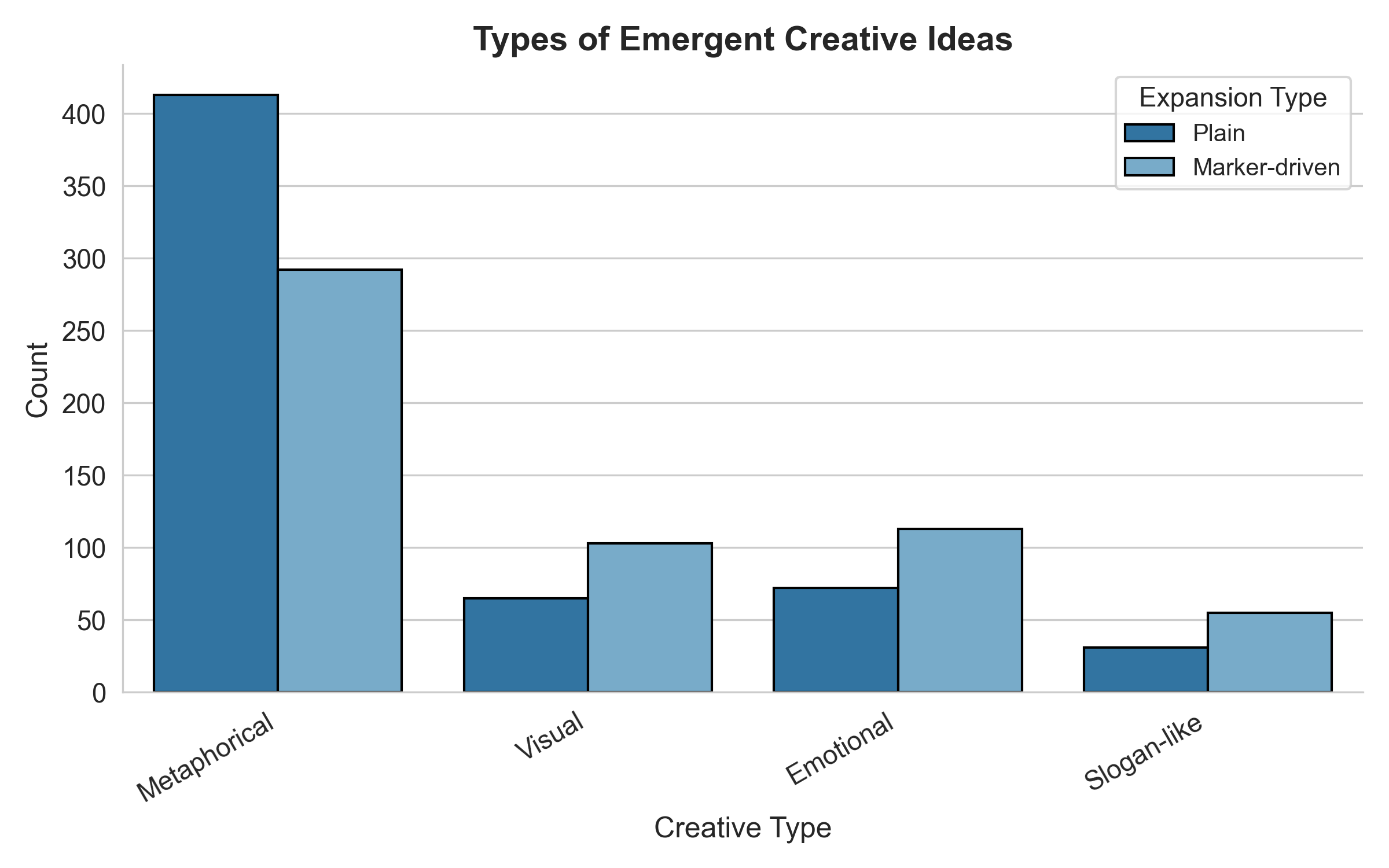}
    \caption{Comparison of emergent creative idea types across Plain and Marker-driven expansions.}
    \label{fig:marker_emergent_idea_types}
\end{figure}

To check whether this trend extended beyond the 100-ad sample, we examined the full set of 1,045 extreme expansions using word- and bigram-frequency analysis. The word cloud in Figure \ref{fig:bigram-wordcloud} highlights how phrases such as "whisper secrets," "cutting edge," "endless possibilities," and "crafted sanctuary" recur across very different ads. This pattern suggested that much of the apparent creativity comes from recycling a narrow pool of high-frequency advertising clichés, raising questions about how "original" these emergent ideas truly are.
We then repeated the qualitative analysis on 100 ads expanded under the marker-driven prompting condition. The difference was clear. As shown in Figure \ref{fig:marker_emergent_idea_types}, the distribution of creative types shifted towards greater balance: metaphorical ideas dropped, while emotional, visual, and slogan-like content increased. In other words, when models were given explicit markers as creative cues, they moved closer to the stylistic mix seen in real advertising ideas, short slogans, emotional hooks, and concrete visual imagery.
\vspace{0.1em}

This showed that LLMs can be steered in new directions when given stronger creative cues. Marker-driven expansions did not remove the model’s habit of repeating familiar phrases, but they did push it closer to the style of real advertising creativity. The key insight is that LLM “creativity” is not fixed. It shifts depending on how we guide the model. Plain expansions revealed its default habit of leaning on metaphors, while marker-driven prompting showed that with structure, the model can diversify, even if it still relies on familiar tropes.

\section{Discussion}

Our study highlights that large language models tend to regress towards generic phrasing when challenged with creative generation, and several factors help explain this outcome. The next-token prediction objective encourages models to reproduce high-probability continuations rather than preserve rare creative features, while training corpora dominated by mainstream patterns further reinforce safe, conventional phrasing. Decoding methods also favor tokens with higher likelihood, which accelerates the loss of metaphors, emotions, and visual cues observed in our contractions. These mechanisms explain why expansions increased surface novelty but failed to restore originality: lexical diversity alone cannot counteract the statistical pull towards the mean. Marker-driven prompting improved alignment because it reintroduced structure that models would otherwise not prioritize. Building on these insights, future work could strengthen creativity recovery by designing training objectives that reward diversity and penalize clichés, by adopting adaptive sampling strategies that adjust when repetition or collapse occurs, and by incorporating richer evaluation measures such as human judgments or prototype-distance scores. Together, these refinements would make studies of creativity more robust and guide the development of models that better preserve originality during creative generation tasks.

\section{Conclusion}
\label{sec:conclusion}

We probed creativity loss and recovery in LLMs via a two-phase stress test on 1{,}045 ad concepts across six models (three runs each). In Phase~1 (forgetting), progressive contraction erased creative detail first, cosine and METEOR fell, lexical entropy dropped, and distinct 4-grams rose; in Phase~2 (expansion), plain regeneration increased length and surface variety but did not restore fidelity, producing outputs that sounded new but relied on familiar clichés, while marker-driven prompting partially recovered alignment. From these results we draw three key insights. First, regression towards generic phrasing is a consistent property across models. Second, surface novelty is not the same as true creativity, since new wording can still hide shallow ideas. Third, external structure such as creative markers can guide models towards better alignment, but they cannot fully restore originality. We interpret this pattern as regression towards a domain prototype, as described by Galton’s Law. Despite repeated runs and full mean$\pm$SD reporting, limitations include domain specificity (ads), metric/LLM-judge dependence, and the information advantage of markers. In short, structure and targeted signals can steer generation away from mediocrity, suggesting a path towards creativity-sensitive LLMs.

\pagebreak

\bibliographystyle{apalike}
\bibliography{references}

\begin{thebibliography}{}

\bibitem[Anderson et~al., 2024]{anderson2024homogenization}
Anderson, B.~R., Shah, J.~H., and Kreminski, M. (2024).
\newblock Homogenization effects of large language models on human creative ideation.
\newblock In {\em Proceedings of the 16th conference on creativity \& cognition}, pages 413--425.

\bibitem[Banerjee and Lavie, 2005]{banerjee-lavie-2005-meteor}
Banerjee, S. and Lavie, A. (2005).
\newblock {METEOR}: An automatic metric for {MT} evaluation with improved correlation with human judgments.
\newblock In Goldstein, J., Lavie, A., Lin, C.-Y., and Voss, C., editors, {\em Proceedings of the {ACL} Workshop on Intrinsic and Extrinsic Evaluation Measures for Machine Translation and/or Summarization}, Ann Arbor, Michigan. Association for Computational Linguistics.

\bibitem[Brown et~al., 2020]{brown2020language}
Brown, T., Mann, B., Ryder, N., Subbiah, M., Kaplan, J.~D., Dhariwal, P., Neelakantan, A., Shyam, P., Sastry, G., Askell, A., et~al. (2020).
\newblock Language models are few-shot learners.
\newblock {\em Advances in neural information processing systems}, 33:1877--1901.

\bibitem[Chen et~al., 2024]{chen2024from}
Chen, Z., Wang, S., Xiao, T., Wang, Y., Chen, S., Cai, X., He, J., and Wang, J. (2024).
\newblock From scaling law to sub-scaling law: Understanding the diminishing returns of larger models.

\bibitem[Chiang and yi~Lee, 2023]{chiang2023largelanguagemodelsalternative}
Chiang, C.-H. and yi~Lee, H. (2023).
\newblock Can large language models be an alternative to human evaluations?

\bibitem[Contagious, 2014]{contagious2014volvo_live_tests}
Contagious (2014).
\newblock Insight \& strategy: Volvo, live tests – the thinking behind the auto manufacturer’s branded content series, including van damme’s epic splits.
\newblock \url{https://www.contagious.com/news-and-views/insight-strategy-live-tests}.

\bibitem[Fou, 2011]{forbes_thinkdifferent}
Fou, A. (2011).
\newblock The real story behind apple's 'think different' campaign.
\newblock {\em Forbes}.
\newblock Accessed: 2025-09-29.

\bibitem[Franceschelli and Musolesi, 2024]{franceschelli2024creativity}
Franceschelli, G. and Musolesi, M. (2024).
\newblock On the creativity of large language models.
\newblock {\em AI \& SOCIETY}, pages 1--11.

\bibitem[Francis~Galton's, 1886]{galton1886regression}
Francis~Galton's, F. (1886).
\newblock Regression towards mediocrity in hereditary stature.
\newblock {\em The Journal of the Anthropological Institute of Great Britain and Ireland}, 15:246--263.

\bibitem[Gu and Krenn, 2024]{gu2024interesting}
Gu, X. and Krenn, M. (2024).
\newblock Interesting scientific idea generation using knowledge graphs and llms: Evaluations with 100 research group leaders.
\newblock {\em arXiv preprint arXiv:2405.17044}.

\bibitem[Gunter et~al., 2024]{gunter2024apple}
Gunter, T., Wang, Z., Wang, C., Pang, R., Narayanan, A., Zhang, A., Zhang, B., Chen, C., Chiu, C.-C., Qiu, D., et~al. (2024).
\newblock Apple intelligence foundation language models.
\newblock {\em arXiv preprint arXiv:2407.21075}.

\bibitem[Haase et~al., 2025]{haase2025has}
Haase, J., Hanel, P.~H., and Pokutta, S. (2025).
\newblock Has the creativity of large-language models peaked? an analysis of inter-and intra-llm variability.
\newblock {\em arXiv preprint arXiv:2504.12320}.

\bibitem[Ismayilzada et~al., 2024]{ismayilzada2024evaluating}
Ismayilzada, M., Stevenson, C., and van~der Plas, L. (2024).
\newblock Evaluating creative short story generation in humans and large language models.
\newblock {\em arXiv preprint arXiv:2411.02316}.

\bibitem[Kumar et~al., 2025]{kumar2025human}
Kumar, H., Vincentius, J., Jordan, E., and Anderson, A. (2025).
\newblock Human creativity in the age of llms: Randomized experiments on divergent and convergent thinking.
\newblock In {\em Proceedings of the 2025 CHI Conference on Human Factors in Computing Systems}, pages 1--18.

\bibitem[Li et~al., 2025]{li2025visual}
Li, Y., Lai, Z., Bao, W., Tan, Z., Dao, A., Sui, K., Shen, J., Liu, D., Liu, H., and Kong, Y. (2025).
\newblock Visual large language models for generalized and specialized applications.
\newblock {\em arXiv preprint arXiv:2501.02765}.

\bibitem[Liu et~al., 2025]{liu2025not}
Liu, E., Bertsch, A., Sutawika, L., Tjuatja, L., Fernandes, P., Marinov, L., Chen, M., Singhal, S., Lawrence, C., Raghunathan, A., et~al. (2025).
\newblock Not-just-scaling laws: Towards a better understanding of the downstream impact of language model design decisions.
\newblock {\em arXiv preprint arXiv:2503.03862}.

\bibitem[Lu et~al., 2024]{lu2024benchmarking}
Lu, Y., Wang, D., Li, T., Jiang, D., Khudanpur, S., Jiang, M., and Khashabi, D. (2024).
\newblock Benchmarking language model creativity: A case study on code generation.
\newblock {\em arXiv preprint arXiv:2407.09007}.

\bibitem[Michael, 2025]{michael2025vwlemon}
Michael, P. (2025).
\newblock Volkswagen’s lemon ad: A marketing masterstroke.
\newblock \url{https://empathyfirstmedia.com/volkswagen-lemon-ad/}.

\bibitem[Mikolov et~al., 2013]{mikolov2013efficient}
Mikolov, T., Chen, K., Corrado, G., and Dean, J. (2013).
\newblock Efficient estimation of word representations in vector space.
\newblock {\em arXiv preprint arXiv:1301.3781}.

\bibitem[Mohammadi, 2024]{mohammadi2024creativity}
Mohammadi, B. (2024).
\newblock Creativity has left the chat: The price of debiasing language models.
\newblock {\em arXiv preprint arXiv:2406.05587}.

\bibitem[Muennighoff et~al., 2023]{muennighoff2023scaling}
Muennighoff, N., Rush, A., Barak, B., Le~Scao, T., Tazi, N., Piktus, A., Pyysalo, S., Wolf, T., and Raffel, C.~A. (2023).
\newblock Scaling data-constrained language models.
\newblock {\em Advances in Neural Information Processing Systems}, 36:50358--50376.

\bibitem[Pasquinelli and Joler, 2021]{pasquinelli2021nooscope}
Pasquinelli, M. and Joler, V. (2021).
\newblock The nooscope manifested: Ai as instrument of knowledge extractivism.
\newblock {\em AI \& society}, 36(4):1263--1280.

\bibitem[Peters, 2024]{embryo_justdoit}
Peters, M. (2024).
\newblock The history of nike’s ‘just do it’ campaign.
\newblock \url{https://embryo.com/blog/history-of-nike-just-do-it/}.
\newblock Published 16 September 2024; last modified 23 December 2024.

\bibitem[Sanburn, 2015]{sanburn2015time_vwforce}
Sanburn, J. (2015).
\newblock The ad that changed super bowl commercials forever.
\newblock {\em Time}.

\bibitem[Team et~al., 2023]{team2023gemini}
Team, G., Anil, R., Borgeaud, S., Alayrac, J.-B., Yu, J., Soricut, R., Schalkwyk, J., Dai, A.~M., Hauth, A., Millican, K., et~al. (2023).
\newblock Gemini: a family of highly capable multimodal models.
\newblock {\em arXiv preprint arXiv:2312.11805}.

\bibitem[Tian et~al., 2024]{tian2024large}
Tian, Y., Huang, T., Liu, M., Jiang, D., Spangher, A., Chen, M., May, J., and Peng, N. (2024).
\newblock Are large language models capable of generating human-level narratives?
\newblock {\em arXiv preprint arXiv:2407.13248}.

\bibitem[Vaswani et~al., 2017]{vaswani2017attention}
Vaswani, A., Shazeer, N., Parmar, N., Uszkoreit, J., Jones, L., Gomez, A.~N., Kaiser, {\L}., and Polosukhin, I. (2017).
\newblock Attention is all you need.
\newblock {\em Advances in neural information processing systems}, 30.

\bibitem[Wenger and Kenett, 2025]{wenger2025we}
Wenger, E. and Kenett, Y. (2025).
\newblock We're different, we're the same: Creative homogeneity across llms.
\newblock {\em arXiv preprint arXiv:2501.19361}.

\bibitem[Zheng et~al., 2023]{zheng2023judgingllmasajudgemtbenchchatbot}
Zheng, L., Chiang, W.-L., Sheng, Y., Zhuang, S., Wu, Z., Zhuang, Y., Lin, Z., Li, Z., Li, D., Xing, E.~P., Zhang, H., Gonzalez, J.~E., and Stoica, I. (2023).
\newblock Judging llm-as-a-judge with mt-bench and chatbot arena.

\end{thebibliography}
 \pagebreak

\begin{appendices}
\renewcommand{\thefigure}{\Alph{section}\arabic{figure}}
\makeatletter
\@addtoreset{figure}{section}
\makeatother

\renewcommand{\thetable}{\Alph{section}\arabic{table}}
\makeatletter
\@addtoreset{table}{section}
\makeatother

  \section{Illustrative Advertising Case Studies}
\label{appendix:a}
\subsection{Volkswagen “Darth Vader” Commercial}
\begin{figure}[H]
    \centering
    \includegraphics[width=0.8\textwidth]{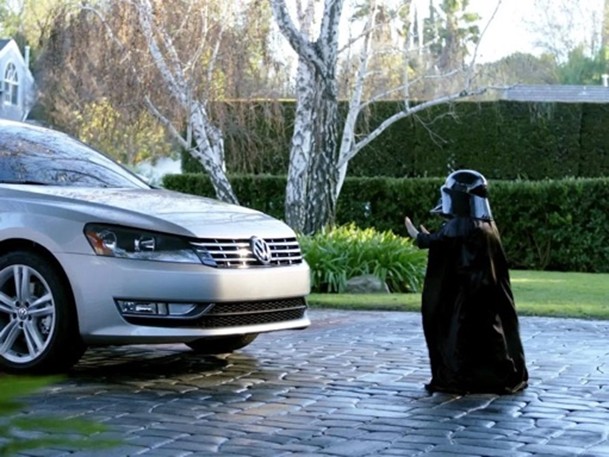}
    \caption{Volkswagen “Darth Vader” Commercial Scene}
    \label{fig:a1}
\end{figure}

This 2011 Super Bowl ad puts a normal family sedan next to a kid dressed as Darth Vader. The child attempts to use “the Force” on various household objects before turning to the Passat; the car’s remote‑start feature is triggered by a parent, creating a moment of surprise and delight. The power of the ad lies in this visual joke and the emotional payoff. Figure \ref{fig:a1} shows still visuals of this specific ad. 

Table \ref{tab:darth-vader} summarises the key elements of the original commercial and notes which parts were reproduced by a vanilla LLM when prompted with a generic brief about a family car with remote start. The percentages are approximate and are for illustration only.
\begin{table}[H]
  \centering
  \begin{tabular}{|p{5cm}|p{2.2cm}|p{7.5cm}|}
    \hline
    \textbf{Script component} & \textbf{LLM coverage} & \textbf{Technical reason for shortfall} \\ \hline
    Child in Darth Vader costume &
    $\sim$50\% &
    Rare combination in the training data; low prior probability in the latent space \\
    Family Volkswagen Passat \& remote start action &
    $\sim$70\% &
    More common automotive descriptions appear in data, though context‐specific cues may vary \\
    Subtle emotional cues (child’s wonder, frustration) &
    $\sim$40\% &
    Lack of fine‐grained affect modelling and long‐range narrative planning \\
    Overall narrative structure \& tension &
    $\sim$60\% &
    LLMs favour safe, average outputs; insufficient modelling of unexpected narrative twists \\
    Integration of visual and textual metaphors (Force, etc.) &
    $\sim$45\% &
    Limited capacity to blend abstract cultural references with concrete action seamlessly \\ \hline
  \end{tabular}
  \caption{Feature Coverage – Darth Vader–Volkswagen Commercial}
  \label{tab:darth-vader}
\end{table}

\subsection{Volvo Trucks “Epic Split” Commercial}
\begin{figure}[H]
    \centering
    \includegraphics[width=0.8\textwidth]{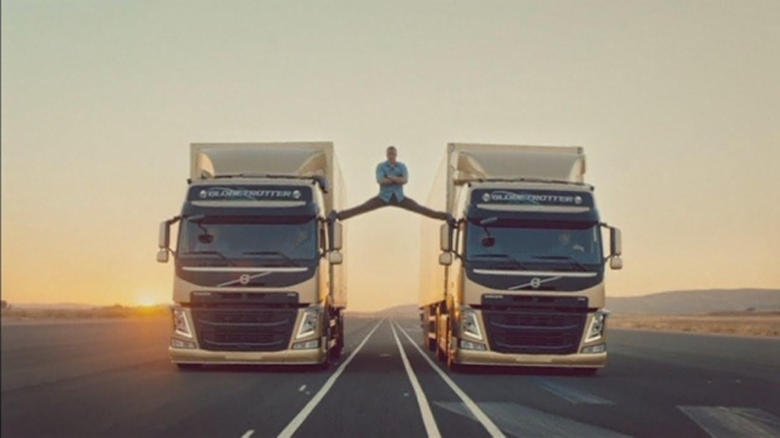}
    \caption{Volvo Trucks “Epic Split” Commercial Scene}
    \label{fig:a2}
\end{figure}

In this 2013 commercial, Jean‑Claude Van Damme performs a split between two moving Volvo trucks to demonstrate their stability. The ad combines technical information about braking and steering systems with a dramatic, memorable stunt. Figure \ref{fig:a2} shows still visuals of this specific ad. 

Table \ref{tab:epic-split}highlights the main narrative beats and shows which ones a generic LLM reproduced when asked to generate a truck‑safety commercial.

\begin{table}[H]
  \centering
  
  \begin{tabular}{|p{5cm}|p{2.2cm}|p{7.5cm}|}
     \hline
    \textbf{Script component} & \textbf{LLM coverage} & 
    \textbf{Technical reason for shortfall} \\ \hline
    Basic Safety Narration (e.g., braking, stability) & $\sim$70\% & High frequency in training data; high token probabilities \\ 
    Dramatic Safety Maneuvers (e.g., epic split‑like actions) & $\sim$50\% & Low probability due to rarity; creative elements underrepresented \\ \
    Technical Engineering Details (specific safety tech) & $\sim$65\% & Moderate probability from technical manuals, but may lack nuance \\ 
    Subtle Emotional and Trust Cues & $\sim$40\% & Limited capacity for fine‑grained affect modeling in highly specific contexts \\ 
    Integration of Visual Metaphors (blending art \& engineering) & $\sim$45\% & Requires cross‑domain knowledge; token‑level mixing not strongly reinforced in training \\ \hline
    
  \end{tabular}
  \caption{Feature Coverage – Volvo Trucks “Epic Split” Commercial}
  \label{tab:epic-split}
\end{table}

\section{Prompts}
This appendix presents the full set of prompts used in our experiments, covering all phases of the study: contraction in Phase 1, plain expansion in Phase 2, and marker-driven expansion as a Phase 2 variant. Each prompt was designed to be clear, consistent, and reproducible, so that the models faced the same conditions across runs. For all generations, we fixed the sampling temperature at 0.7 to strike a balance between stability and allowing room for creative variation.
\label{appendix:b}

\subsection{Phase 1: Forgetting}

\begin{figure}[H]
  \centering
  \begin{adjustbox}{max width=\linewidth, max height=.85\textheight}
    \begin{tcolorbox}[
      breakable,
      title={Contraction Prompt},
      colback=white, colframe=black,
      colbacktitle=black, coltitle=white,
      fonttitle=\bfseries, boxrule=0.8pt, arc=3pt,
      left=6pt,right=6pt,top=6pt,bottom=6pt
    ]
\begin{lstlisting}[language=Python, basicstyle=\ttfamily\scriptsize, breaklines=true,
  columns=fullflexible, keepspaces=true, showstringspaces=false, upquote=true]

def build_prompt(concept, compression_level):
    base = "Your task is to contract this advertisement concept to the specified level of contraction."

    level_instructions = {
        "mild": (
            "Remove about 35% of the details. Keep most of the structure."
            "Provide ONLY the contracted version without any explanations."
        ),
        "moderate": (
            "Remove about 70% of the details. The result should be much briefer but still clear."
            "Provide ONLY the contracted version without any explanations."
        ),
        "extreme": (
            "Remove about 95% of the details. Leave ONLY 1-2 sentences."
            "Provide ONLY the contracted version without any explanations."
        )
    }
    instr = level_instructions[compression_level]

    return (
        f"{base}"
        f"Contraction level: {compression_level.upper()}"
        f"{instr}"
        f"Original ad concept: {concept}"
        f"Contracted version:"
    )
\end{lstlisting}
    \end{tcolorbox}
  \end{adjustbox}
  \caption{Prompt used in Phase 1 (Forgetting) to contract advertisement concepts.}
  \label{fig:contraction-prompt}
\end{figure}

\subsection{Phase 2: Plain Expansion Prompt}

\begin{figure}[H]
  \centering
  \begin{adjustbox}{max width=\linewidth, max height=.85\textheight}
    \begin{tcolorbox}[
      title={Expansion Prompt},
      colback=white, colframe=black,
      colbacktitle=black, coltitle=white,
      fonttitle=\bfseries, boxrule=0.8pt, arc=3pt,
      left=4pt,right=4pt,top=4pt,bottom=4pt
    ]
\begin{lstlisting}[language=Python, basicstyle=\ttfamily\tiny, breaklines=true,
  columns=fullflexible, keepspaces=true, showstringspaces=false, upquote=true]

def build_expansion_prompt(input_text, level, original_word_count):
    # Calculate target word counts per level
    word_targets = {
        "mild": int(original_word_count * 0.3),      # ~30% of original
        "moderate": int(original_word_count * 0.6),  # ~60% of original
        "extreme": original_word_count               # Full length
    }

    instructions = {
        "mild": (
            "Add 1-2 simple, meaningful details. Maintain the core message. "
            "Keep it concise - around 1-2 sentences. Try to stay brief and direct. "
            "Provide ONLY the rewritten version without any explanations or preamble."
        ),
        "moderate": (
            "Build on the previous version with 2-3 sentences of engaging or descriptive elements. "
            "Introduce a fresh perspective or unexpected twist while staying close to the original meaning. "
            "Provide ONLY the rewritten version without any explanations or preamble."
        ),
        "extreme": (
            "Build on the previous version with 3-4 imaginative sentences. Add surprising or expressive details. "
            "Use lateral thinking or vivid storytelling while preserving the original intent. "
            "Provide ONLY the rewritten version without any explanations or preamble."
        )
    }
    target_words = word_targets[level]

    return f'''
You are a creative assistant.
Your task is to expand the following advertisement while keeping its core idea intact.
Expansion Level: {level.upper()}
Instructions: {instructions[level]}
The original ad was approximately {original_word_count} words long. Aim for approximately {target_words} words.
Do not significantly exceed this length.'''
\end{lstlisting}
    \end{tcolorbox}
  \end{adjustbox}
  \caption{Prompt used in Phase 2 (Expansion) to regenerate advertisement concepts.}
  \label{fig:expansion-prompt}
\end{figure}

\vspace{10 em}
\subsection{Phase 2: Marker Driven Expansion Prompt}
\vspace{0.5 em}

\begin{figure}[H]
  \centering
  \begin{adjustbox}{max width=\textwidth, max height=.85\textheight}
    \begin{tcolorbox}[
      title={Marker-driven Prompt},
      colback=white,
      colframe=black,
      colbacktitle=black,
      coltitle=white,
      fonttitle=\bfseries,
      boxrule=0.8pt,
      arc=3pt,
      left=4pt,right=4pt,top=4pt,bottom=9pt
    ]
\begin{lstlisting}[language=Python, basicstyle=\ttfamily\tiny, breaklines=true,
  columns=fullflexible, keepspaces=true, showstringspaces=false, upquote=true]

def build_marker_prompt(input_text, markers, level, original_word_count):
    """Marker-driven prompt: mirrors plain prompt + explicit creative markers"""

    # Target length (same logic)
    word_targets = {
        "mild": int(original_word_count * 0.3),      # ~30% of original
        "moderate": int(original_word_count * 0.6),  # ~60% of original
        "extreme": original_word_count               # Full length
    }

    # Instructions (same tone, but add marker guidance)
    instructions = {
        "mild": (
            "Add 1-2 simple, meaningful details. Maintain the core message. "
            "Include 1-2 of the most essential creative markers. "
            "Keep it concise - around 1-2 sentences. Try to stay brief and direct. "
            "Provide ONLY the rewritten version without any explanations or preamble."
        ),
        "moderate": (
            "Build on the previous version with 2-3 sentences of engaging or descriptive elements. "
            "Introduce a fresh perspective or unexpected twist while still reflecting the core message. "
            "Include 2-3 creative markers naturally in the flow. "
            "Provide ONLY the rewritten version without any explanations or preamble."
        ),
        "extreme": (
            "Build on the previous version with 3-4 imaginative sentences. "
            "Add surprising or expressive details. Use lateral thinking or vivid storytelling. "
            "Ensure that all the provided creative markers are included meaningfully. "
            "Provide ONLY the rewritten version without any explanations or preamble."
        )
    }

    return f'''
You are a creative assistant.

Your task is to expand the following advertisement while keeping its core idea intact.

Expansion Level: {level.upper()}
Instructions: {instructions[level]}
The original ad was approximately {original_word_count} words long. Do not significantly exceed this length.

Please incorporate the following creative markers into your expansion. 

These represent the most essential ideas from the original ad and must be preserved (you may paraphrase them, but the meaning must stay intact):

Creative Markers:
{marker_text}

Input:
"{input_text}"

Expanded version:
'''.strip()
\end{lstlisting}
    \end{tcolorbox}
  \end{adjustbox}
  \caption{Prompt used in Phase 2 (Marker-Driven Expansion) to regenerate concepts with creative markers.}
  \label{fig:marker-prompt}
\end{figure}

\vspace{0.3 em}

\section{}
\label{appendix:c}
\subsection{Original Advertising Ideas}
This appendix includes seven sample advertisement concepts from the 55mV Idea Database to show the type of material used in our study. These examples capture the creative qualities that made the database a strong foundation for our experiments, including the use of metaphors, emotional tone, vivid visual imagery, and concise, memorable phrasing. Together, they highlight the richness and variety of ideas that we used as a baseline for testing how language models contract and regenerate creative content.
\label{appendix:c1}

\begin{figure}[H]
\centering
\begin{tcolorbox}[colback=white, colframe=black,
  fonttitle=\bfseries, colbacktitle=black, coltitle=white,
  sharp corners, boxrule=0.6pt, title=Original Ideas]
\textbf{Ad-001:} SolaceSense is a new type of air freshener that utilises the power of AI to detect when someone has had a bad day and plays a comforting scent. For example, if the AI detects that you have had a stressful day at work, it will release a calming lavender scent. If it senses that you are feeling under the weather, it will release a refreshing eucalyptus scent.

\vspace{0.3em}

\textbf{Ad-002:} A pharmacy brand that has a range of vitamins and supplements to help keep your skin glowing from the inside out. Print and social campaign that has a modern take on the 'you are what you eat' idea. But instead 'you are what you absorb' showing people with different vitamins and supplement shapes cut out of their bodies like paper dolls.

\vspace{0.3em}

\textbf{Ad-003:} This is a campaign for a tea brand, where we create a series of short films of people in everyday situations having a moment of relaxation by themselves. We are kept at a distance and simply observe the scene. We can hear the sounds and the world around them. As the person takes a sip of the tea, we hear their body relax. And then we hear an amplified version of the sounds of the body relaxing. It's incredibly satisfying to listen to. We hear the heart slow down, the breath deepen, the stomach soften, the muscles relax. We are transported to a world where we can really hear what it sounds like to relax. We end on a super of the tea brand and the line: There's comfort in every cup.

\vspace{0.3em}

\textbf{Ad-004:} A film about the joy of finding the perfect pair of glasses. We see a man looking in a mirror, trying on different pairs. He is in a shop, and the shop assistant is bringing him more and more pairs to try. The man is transformed with each pair from serious and studious, to fun and funky. We see him going about his day, wearing different pairs each time he puts on a new pair, his whole mood changes. He walks out of the shop, but something is missing. He tries on another pair, but still doesn’t feel right. He walks past a poster for [brand] glasses and stops to look. The glasses on the poster are perfect. He walks back to the shop and picks up the pair he saw in the poster. These are the ones. As soon as he puts them on, he feels like himself. He walks out of the shop, now confident and happy. Titles: Find the perfect frames for you.

\vspace{0.3em}

\textbf{Ad-005:} We see a series of people who are in various emotional states. The people appear in a frame in a gallery. As they stand there in their emotional state a dog enters the frame and sits down next to them. The person then begins to pat the dog and we see them visibly relax and experience a sense of calm. Each person is matched to a dog that has the same emotional state.
\vspace{0.3em}

\textbf{Ad-006:} We all take water for granted. This film is about reminding people how important water is to us. We show a series of everyday moments of people experiencing water. A woman in the shower. A child playing in a bath. A man drinking a glass of water. A family barbecuing in the backyard. A group of kids playing with a garden hose. The moments are all shot in slow motion, with a beautiful, dreamy quality. Over the top of the images, a poem is read. The poem is about the beauty and power of water. The film ends with a super: Sydney Water. Water is life.
\vspace{0.3 em}

\textbf{Ad-007:} We see a range of kids doing a range of adult activities that we can't quite believe they are doing. They are dressed like adults and carrying out the respective tasks. One kid is sitting in a cafe with another adult and they are discussing politics, another is leading a boardroom meeting, another is doing an interview for a job, another is giving a lecture. They all end with the adult person saying 'you're amazing thank you so much.' The kid then goes into their bag and pulls out a toy that they hand over to the adult. VO: Bring the kid back into your kid. A store that sells second hand toys.
\end{tcolorbox}
\caption{Examples of original advertisement ideas used as input for the experiments.}
\label{fig:original-ads}
\end{figure}

\vspace{5 em}

\subsection{Phase 1- Illustrative Example}

To illustrate how the contraction process operates, we provide an example of an advertisement idea reduced across the three levels. At the mild stage, the text is shortened but still retains most of its creative details and tone. At the moderate stage, much of the metaphorical and emotional content is stripped away, leaving a simpler and more straightforward version. At the extreme stage, the idea is compressed into one sentences that capture only the most basic product message. This progression shows how creative richness gradually fades as the level of contraction increases.
\vspace{3 em}

\begin{figure}[H]
  \centering
  \begin{tcolorbox}[colback=white,colframe=black,title=Phase 1 Forgetting Example]
    \textbf{Original:} Tree people, people that look like trees and are camouflaged, hidden in plain sight. 
    When they suddenly move they scare people. We see a series of tree people hiding in and around cars 
    and when kids, cyclists, people walk past them they suddenly move and scare them. 
    The idea being that often we can't see everything around us but the car sensors can." \\[0.1em]

    \textbf{Mild:} Tree people, camouflaged to look like trees, hide in plain sight. 
    They suddenly move to scare passersby. A series of tree people surprise kids, cyclists, 
    and pedestrians near cars, highlighting that while we can't always see everything, car sensors can. \\[0.1em]

    \textbf{Moderate:} Camouflaged tree people suddenly move and scare passersby, 
    highlighting that car sensors detect hidden things we can't see. \\[0.1em]

    \textbf{Extreme:} Car sensors detect what you can't see.
  \end{tcolorbox}
  \caption{Phase 1 forgetting example}
  \label{fig:phase1_forgetting_example}
\end{figure}

\vspace{3 em}
\subsection{Phase2 plain expansion Illustrative example}
\label{appendix:c3}
To illustrate the plain expansion process, we present an example drawn from the VitaFresh Organic Produce campaign. The original advertisement is shown here only to provide context on the underlying concept; in the actual experiment, the model has no access to the original during expansion. Instead, the extreme contracted form serves as the sole input. At the mild level, the model adds a few descriptive details to the tagline. At the moderate level, the ad develops into a short narrative that frames eating as an art and highlights freshness as a sensory experience. At the extreme level, the model produces an elaborate, imaginative version that emphasizes imagery, metaphor, and emotional tone. This stepwise progression demonstrates how, without additional guidance, the model attempts to restore creative richness when working from a highly compressed starting point.

\begin{figure}[H]
\centering
\begin{tcolorbox}[colback=white, colframe=black,
  title=Plain Expansion Example,
  colbacktitle=black, coltitle=white,
  fonttitle=\bfseries,
  sharp corners, boxrule=0.6pt,
  width=0.95\textwidth]

\textbf{Original:} A brand campaign for VitaFresh Organic Produce. In the world of food, there are two types of people: the people who eat to live, and the people who live to eat. VitaFresh Organic Produce is for the latter. The people who eat to live, they do it out of necessity, they don't care too much, they aren't passionate about it. But the people who live to eat, they care about what they eat, they are passionate about food, they savour every bite. This campaign is for the people who live to eat. With the tagline: \textit{Made for the people who live to eat.} \\[0.5em]

\textbf{Base Concept:} \textit{VitaFresh Organic Produce: Made for those who live to eat.} \\[0.5em]

\textbf{Mild Expansion:} \textit{VitaFresh Organic Produce: Made for those who live to eat and savor every bite of nature’s freshest flavors.} \\[0.5em]

\textbf{Moderate Expansion:} Discover VitaFresh Organic Produce—crafted for those who see eating as an art and treasure every taste. More than food, it is a celebration of the earth’s bounty, grown and harvested with care to bring pure delight to your table. Each bite delivers freshness that turns an ordinary meal into a feast for the senses. \\[0.5em]

\textbf{Extreme Expansion:} Step into the world of VitaFresh Organic Produce, where every bite is a symphony of flavors for those who savor the art of eating. Picture an orchard at dawn, dew glistening on sun-kissed fruit, each piece a gift of nature’s generosity. Our produce is not just nourishment; it is an ode to the land, cultivated with devotion to elevate your culinary journey. Every mouthful bursts with freshness, transforming meals into a celebration of the planet’s purest offerings.

\end{tcolorbox}
\caption{Phase 2 plain expansion example}
\label{fig:phase2_plain_expansion_example}
\end{figure}

\vspace{3 em}
\subsection{Phase 2-Marker-Driven Expansion Illustrative Example}
\label{appendix:c4}
To illustrate the marker-driven expansion process, we include an example where the model expands an ad concept starting from its extreme contracted form. The original advertisement is shown only to indicate the underlying idea; during the experiment, the model has no access to it. Instead, it is provided with a small set of creative markers extracted from the original, covering different categories such as visual, emotional, brand-specific, and slogan-like cues. These markers are introduced progressively across mild, moderate, and extreme levels of expansion. The example highlights how this additional guidance enables the model to generate richer, more expressive outputs compared to plain expansion.
\begin{figure}[H]
\centering
\begin{tcolorbox}[colback=white, colframe=black,
  title=Marker-Driven Expansion Example,
  colbacktitle=black, coltitle=white,
  fonttitle=\bfseries,
  sharp corners, boxrule=0.6pt,
  width=0.95\textwidth] 

\textbf{Original:} A film for a free-range chicken brand who wants to celebrate how they raise their chickens. The film is a beautifully shot, slow film of the inside of a chicken shed, which is actually a sprawling, open field, with trees, and sunshine. As the camera slowly moves through the shed, we see free-range chickens everywhere. They are clucking and pecking, and there are no cages in sight. We hear the sound of the birds, and the wind in the trees. There are no people in sight. The film ends with a super that says ``Our chickens live free range, so you can too.'' \\[0.5em]

\textbf{Extreme Input (Base Concept):} ``Free-range chickens roam freely in open fields. `Our chickens live free range, so you can too.' '' \\[0.5em]

\textbf{Markers:} \\
-- Chicken shed: a sprawling field of trees under sunshine. (Visual) \\
-- We see free-range chickens everywhere. (Brand-specific) \\
-- Hearing birdsong and wind rustling through trees. (Emotional) \\
-- Our chickens live free range, so you can too. (Slogan-like) \\[0.5em]

\textbf{Mild Expansion:} In a sprawling field of sunlit trees, free-range chickens roam happily, accompanied by the gentle birdsong and rustling leaves. Our chickens live free range, so you can too. \\[0.5em]

\textbf{Moderate Expansion:} Beneath a canopy of sunlight filtering through lush trees, our free-range chickens wander freely, creating a symphony with the gentle birdsong and the whisper of wind through leaves. Picture a place where nature thrives, and every step is a dance of freedom. As our chickens relish their open world, you too can embrace the spirit of liberation. Experience the joy of knowing that our chickens live free range, so you can savor the taste of freedom in every bite. \\[0.5em]

\textbf{Extreme Expansion:} In a magical realm where sunlight weaves through a tapestry of lush, whispering trees, our free-range chickens roam with a grace that rivals the wind itself. Picture a scene where every chicken, a feathered troubadour, adds its own notes to the symphony of nature's orchestra, harmonizing with the melodious birdsong and the gentle rustle of leaves. This is not just a farm; it's a haven where the spirit of freedom dances across the landscape. As our chickens revel in their boundless paradise, you, too, can taste the essence of liberation. Relish in the fact that our chickens live free range, so you can indulge in the true flavor of freedom.

\end{tcolorbox}
\caption{Phase 2 marker-driven expansion example}
\label{fig:phase2_marker_expansion_example}
\end{figure}

\vspace{1em}

\subsection{List of creative Markers Extracted for Marker-driven version}
\label{appendix:c5}
The creative markers were extracted directly from the original advertisement ideas. We looked for distinctive elements such as vivid imagery, emotional expressions, metaphors, short slogan-like phrases, and brand-specific references. Each marker was selected to capture the most creative part of the idea while avoiding generic or filler language. These markers were then used in the marker-driven expansion stage to guide the models and help them regenerate outputs that retained richer creative qualities.  To provide a clearer view of the guidance given to the models, below is a sample of the creative markers extracted during Phase 2 marker-driven expansion. These markers below are randomly chosen from a subset of advertisements and are shown together with their assigned categories, visual, emotional, metaphorical, slogan-like, and brand-specific. This illustrative list highlights the range and diversity of cues used to support the regeneration process.
\vspace{2 em}

\begin{figure}[H]
\centering
\begin{tcolorbox}[
  title={Sample Creative Markers},
  colback=white, colframe=black,
  colbacktitle=black, coltitle=white,
  fonttitle=\bfseries, boxrule=0.8pt, arc=3pt,
  left=4pt,right=4pt,top=4pt,bottom=4pt
]

\begin{itemize}
    \item Diverse ages and backgrounds united in cooking eggs. \textit{(Visual)}
    \item They all crack the egg at the same time. \textit{(Visual)}
    \item The sound is a “crunchy” but satisfying sound. \textit{(Emotional)}
    \item The visual treatment is modern with a pop of colour. \textit{(Visual)}
    \item 18-year-old girl going to a concert. \textit{(Visual)}
    \item Wearing the same outfit as her 50-year-old mother. \textit{(Visual)}
    \item Same scene with a 70-year-old grandmother. \textit{(Visual)}
    \item No matter what your age you can wear this brand. \textit{(Brand-specific)}
    \item To see which one can be eaten the quickest. \textit{(Brand-specific)}
    \item A fire extinguisher with a camera. \textit{(Visual)}
    \item A vacuum cleaner that learns all the swear words. \textit{(Emotional)}
    \item A thermostat that goes to 11. \textit{(Metaphorical)}
    \item A smart bed that tells you to go to sleep. \textit{(Brand-specific)}
    \item Quick sleepers and night owls. \textit{(Emotional)}
    \item Night People. This one’s for you. \textit{(Slogan-like)}
    \item Scan QR code for a playlist of soothing sleep sounds. \textit{(Brand-specific)}
    \item A tree that is carved with the full body of a jaguar. \textit{(Visual)}
    \item Lights are flashing as if it were a nightclub. \textit{(Metaphorical)}
    \item The person is shocked and closes the door. \textit{(Emotional)}
    \item The pocket that can save your life. \textit{(Slogan-like)}
\end{itemize}
\end{tcolorbox}
    \caption{Sample list of creative markers extracted in Phase 2 marker-driven expansion. Each marker is shown with its category label.}
    \label{fig:marker-sample}
\end{figure}
\pagebreak

\section{}
\label{appendix:d}

\subsection{Metric Computation}
This appendix describes the procedures used to compute the four evaluation metrics reported in Phases 1 and 2. Each metric was applied at the advertisement level and then averaged across the 1,045 ads in the dataset. 
\vspace{1 em}

\begin{itemize}
    \item \textbf{Cosine Similarity.}  
    Semantic alignment between the regenerated text and the original advertisement was measured using TF--IDF weighted representations. Each text pair was vectorized with scikit-learn's \texttt{TfidfVectorizer}, and cosine similarity was computed as:
    \[
    \text{cosine}(x,y) = \frac{x \cdot y}{\|x\|\|y\|}
    \]
    where $x$ and $y$ are the TF--IDF vectors. Scores range between 0 and 1, with higher values indicating stronger semantic similarity.
\vspace{2em}

    \item\textbf{METEOR}
    Surface-level correspondence was assessed with METEOR,
    which accounts for synonyms and word-order flexibility. The original advertisement served as the reference, and the regenerated ad as the hypothesis. Both texts were tokenized into words, and the
    algorithm aligned them using exact matches, stemmed matches, and synonym matches. The METEOR score combines precision (the fraction of generated words that overlap with the reference) and recall
    (the fraction of the reference captured in the generated text), with a penalty for fragmented word order. We used the implementation in NLTK's \texttt{meteor\_score.single\_meteor\_score}
    with default parameters.
\vspace{2em}

    \item \textbf{Entropy.}  
    Lexical diversity was estimated using Shannon entropy over the unigram distribution of each text. Word counts were collected using Python's \texttt{Counter}, and entropy was computed as:
    \[
    H = - \sum_{w \in V} p(w) \log_2 p(w)
    \]
    with $p(w)$ the relative frequency of word $w$. Lower values indicate repetition and simplified language, while higher values indicate richer and more varied expression.
\vspace{2em}
    
    \item \textbf{N-gram Uniqueness.}  
    To quantify structural novelty, both texts were tokenized and segmented into sequential 4-grams using NLTK. Two sets of 4-grams were then created, one for the original and one for the generated text. The overlap between these sets was compared against their union, and uniqueness was defined as:
    \[
    U = 1 - \frac{|N_{gen} \cap N_{orig}|}{|N_{gen} \cup N_{orig}|}
    \]
    where $N_{gen}$ and $N_{orig}$ are the sets of 4-grams in the generated and original texts. A higher value indicates greater lexical novelty, while a lower value suggests reuse of original phrasing.
\end{itemize}
\vspace{20 em}

\subsection{Phase 1: Quantitative Stability Across Runs}
Table D1 presents the mean and standard deviation values for Phase 1, calculated across three independent runs of the experiment. The deviations remain consistently low, typically below 0.02, indicating that the outputs produced by the models are highly stable across repetitions. This stability is important because it rules out the possibility that the observed patterns in forgetting are artifacts of random sampling or chance variations in generation. Instead, the results reflect systematic tendencies of the models themselves when asked to contract advertisement ideas. The reliability across runs strengthens the validity of our analysis and provides confidence that the trends reported in the main text are reproducible.
\label{appendix:d1}
\begin{table}[H]
\centering
\small
\begin{tabular}{llcccc}
\toprule
\textbf{Model} & \textbf{Level} & \textbf{Cosine} & \textbf{METEOR} & \textbf{Entropy} & \textbf{N-gram} \\
\midrule
GPT-4o           & Original & 1.000$\pm$0.000 & 1.000$\pm$0.000 & 5.308$\pm$0.000 & 0.000$\pm$0.000 \\
                & Mild     & 0.610$\pm$0.006 & 0.464$\pm$0.005 & 4.909$\pm$0.030 & 0.907$\pm$0.001 \\
                & Moderate & 0.375$\pm$0.004 & 0.222$\pm$0.001 & 4.157$\pm$0.022 & 0.975$\pm$0.000 \\
                & Extreme  & 0.264$\pm$0.005 & 0.126$\pm$0.003 & 3.338$\pm$0.021 & 0.989$\pm$0.000 \\
\midrule
GPT-5 main      & Original & 1.000$\pm$0.000 & 1.000$\pm$0.000 & 5.308$\pm$0.000 & 0.000$\pm$0.000 \\
                & Mild     & 0.663$\pm$0.004 & 0.507$\pm$0.004 & 5.036$\pm$0.002 & 0.882$\pm$0.002 \\
                & Moderate & 0.399$\pm$0.001 & 0.237$\pm$0.002 & 4.373$\pm$0.006 & 0.977$\pm$0.001 \\
                & Extreme  & 0.273$\pm$0.004 & 0.140$\pm$0.001 & 3.668$\pm$0.006 & 0.991$\pm$0.000 \\
\midrule
GPT-5 Thinking  & Original & 1.000$\pm$0.000 & 1.000$\pm$0.000 & 5.308$\pm$0.000 & 0.000$\pm$0.000 \\
                & Mild     & 0.600$\pm$0.002 & 0.457$\pm$0.002 & 4.996$\pm$0.002 & 0.916$\pm$0.001 \\
                & Moderate & 0.380$\pm$0.001 & 0.241$\pm$0.002 & 4.426$\pm$0.009 & 0.980$\pm$0.002 \\
                & Extreme  & 0.326$\pm$0.002 & 0.203$\pm$0.001 & 4.198$\pm$0.009 & 0.985$\pm$0.001 \\
\midrule
GPT-oss-120b    & Original & 1.000$\pm$0.000 & 1.000$\pm$0.000 & 5.308$\pm$0.000 & 0.000$\pm$0.000 \\
                & Mild     & 0.516$\pm$0.001 & 0.361$\pm$0.001 & 4.858$\pm$0.003 & 0.959$\pm$0.001 \\
                & Moderate & 0.390$\pm$0.002 & 0.232$\pm$0.002 & 4.400$\pm$0.009 & 0.982$\pm$0.001 \\
                & Extreme  & 0.323$\pm$0.001 & 0.186$\pm$0.002 & 4.054$\pm$0.005 & 0.986$\pm$0.000 \\
\midrule
Gemini-2.5-pro  & Original & 1.000$\pm$0.000 & 1.000$\pm$0.000 & 5.308$\pm$0.000 & 0.000$\pm$0.000 \\
                & Mild     & 0.733$\pm$0.004 & 0.585$\pm$0.004 & 5.048$\pm$0.006 & 0.818$\pm$0.003 \\
                & Moderate & 0.394$\pm$0.003 & 0.210$\pm$0.001 & 4.177$\pm$0.006 & 0.975$\pm$0.000 \\
                & Extreme  & 0.263$\pm$0.001 & 0.125$\pm$0.002 & 3.580$\pm$0.008 & 0.991$\pm$0.000 \\
\midrule
Claude Sonnet-4 & Original & 1.000$\pm$0.000 & 1.000$\pm$0.000 & 5.308$\pm$0.000 & 0.000$\pm$0.000 \\
                & Mild     & 0.637$\pm$0.014 & 0.489$\pm$0.011 & 4.913$\pm$0.067 & 0.873$\pm$0.002 \\
                & Moderate & 0.388$\pm$0.006 & 0.242$\pm$0.004 & 4.326$\pm$0.046 & 0.974$\pm$0.001 \\
                & Extreme  & 0.286$\pm$0.008 & 0.157$\pm$0.004 & 3.778$\pm$0.043 & 0.987$\pm$0.001 \\
\bottomrule
\end{tabular}
\caption{Phase 1 results across 3 runs per model. Values are reported as mean $\pm$ SD (between-run).}
\label{tab:phase1_stability}
\end{table}
\vspace{5 em}

\subsection{Phase 2: Plain Expansion Quantitative Stability Across Runs}
\label{appendix:d2}
Table D2 reports the plain expansion results averaged across three independent runs for each model. The standard deviations remain consistently low, generally below 0.02, indicating that the patterns observed in expansion are robust and not the product of sampling variance. This consistency shows that, even when starting from highly contracted inputs, the models reproduce similar trajectories of expansion across runs. Such stability strengthens confidence that the findings reflect systematic model behavior rather than random fluctuations, reinforcing the reliability of the trends highlighted in the main text.

\begin{table}[H]
\centering
\small
\begin{tabular}{llcccc}
\toprule
\textbf{Model} & \textbf{Level} & \textbf{Cosine} & \textbf{METEOR} & \textbf{Entropy} & \textbf{N-gram} \\
\midrule
GPT-4o           & Original & 1.000$\pm$0.000 & 1.000$\pm$0.000 & 5.308$\pm$0.000 & 0.000$\pm$0.000 \\
                 & Mild     & 0.237$\pm$0.001 & 0.103$\pm$0.001 & 4.301$\pm$0.008 & 0.988$\pm$0.000 \\
                 & Moderate & 0.270$\pm$0.001 & 0.152$\pm$0.001 & 5.522$\pm$0.009 & 0.997$\pm$0.000 \\
                 & Extreme  & 0.289$\pm$0.001 & 0.165$\pm$0.000 & 6.069$\pm$0.007 & 0.999$\pm$0.000 \\
\midrule
GPT-5 main       & Original & 1.000$\pm$0.000 & 1.000$\pm$0.000 & 5.308$\pm$0.000 & 0.000$\pm$0.000 \\
                 & Mild     & 0.257$\pm$0.002 & 0.111$\pm$0.002 & 4.457$\pm$0.008 & 0.991$\pm$0.000 \\
                 & Moderate & 0.276$\pm$0.001 & 0.158$\pm$0.000 & 5.531$\pm$0.004 & 0.997$\pm$0.000 \\
                 & Extreme  & 0.270$\pm$0.001 & 0.163$\pm$0.002 & 5.966$\pm$0.005 & 0.999$\pm$0.000 \\
\midrule
GPT-5 Thinking   & Original & 1.000$\pm$0.000 & 1.000$\pm$0.000 & 5.308$\pm$0.000 & 0.000$\pm$0.000 \\
                 & Mild     & 0.288$\pm$0.001 & 0.160$\pm$0.000 & 4.876$\pm$0.001 & 0.987$\pm$0.000 \\
                 & Moderate & 0.297$\pm$0.001 & 0.188$\pm$0.001 & 5.716$\pm$0.003 & 0.994$\pm$0.000 \\
                 & Extreme  & 0.291$\pm$0.001 & 0.188$\pm$0.001 & 6.187$\pm$0.007 & 0.997$\pm$0.000 \\
\midrule
GPT-oss-120b     & Original & 1.000$\pm$0.000 & 1.000$\pm$0.000 & 5.308$\pm$0.000 & 0.000$\pm$0.000 \\
                 & Mild     & 0.286$\pm$0.001 & 0.148$\pm$0.000 & 4.705$\pm$0.006 & 0.988$\pm$0.000 \\
                 & Moderate & 0.304$\pm$0.000 & 0.178$\pm$0.001 & 5.639$\pm$0.010 & 0.996$\pm$0.000 \\
                 & Extreme  & 0.289$\pm$0.002 & 0.171$\pm$0.001 & 5.907$\pm$0.023 & 0.993$\pm$0.003 \\
\midrule
Gemini-2.5-pro   & Original & 1.000$\pm$0.000 & 1.000$\pm$0.000 & 5.308$\pm$0.000 & 0.000$\pm$0.000 \\
                 & Mild     & 0.246$\pm$0.000 & 0.118$\pm$0.000 & 4.522$\pm$0.013 & 0.988$\pm$0.002 \\
                 & Moderate & 0.263$\pm$0.003 & 0.146$\pm$0.002 & 5.496$\pm$0.019 & 0.996$\pm$0.001 \\
                 & Extreme  & 0.253$\pm$0.004 & 0.138$\pm$0.003 & 5.667$\pm$0.061 & 0.996$\pm$0.002 \\
\midrule
Claude Sonnet-4  & Original & 1.000$\pm$0.000 & 1.000$\pm$0.000 & 5.308$\pm$0.000 & 0.000$\pm$0.000 \\
                 & Mild     & 0.264$\pm$0.000 & 0.151$\pm$0.001 & 4.894$\pm$0.007 & 0.987$\pm$0.000 \\
                 & Moderate & 0.268$\pm$0.001 & 0.176$\pm$0.002 & 5.843$\pm$0.004 & 0.995$\pm$0.000 \\
                 & Extreme  & 0.216$\pm$0.002 & 0.144$\pm$0.000 & 6.113$\pm$0.006 & 0.998$\pm$0.000 \\
\bottomrule
\end{tabular}
\caption{Phase 2 Plain Expansion results across 3 runs per model. Values are reported as mean $\pm$ SD (between-run).}
\label{tab:phase2_plain_stability}
\end{table}
\vspace{5 em}

\subsection{Phase 2 Marker-Driven:} Table D3 presents the marker-driven expansion results averaged across three independent runs for each model. The standard deviations remain small, typically below 0.02, demonstrating that the outputs are stable and reproducible across runs. This consistency indicates that the observed effects of marker guidance are systematic rather than artifacts of random variation. The stability of these results reinforces the validity of comparing marker-driven expansion against plain expansion, since both display reliable behavior across repeated trials.

\begin{table}[H]
\centering
\small
\begin{tabular}{llcccc}
\toprule
\textbf{Model} & \textbf{Level} & \textbf{Cosine} & \textbf{METEOR} & \textbf{Entropy} & \textbf{N-gram} \\
\midrule
GPT-4o           & Original & 1.000$\pm$0.000 & 1.000$\pm$0.000 & 5.308$\pm$0.000 & 0.000$\pm$0.000 \\
                 & Mild     & 0.405$\pm$0.010 & 0.236$\pm$0.008 & 4.778$\pm$0.041 & 0.932$\pm$0.004 \\
                 & Moderate & 0.359$\pm$0.010 & 0.230$\pm$0.011 & 5.542$\pm$0.087 & 0.983$\pm$0.001 \\
                 & Extreme  & 0.346$\pm$0.013 & 0.218$\pm$0.011 & 6.015$\pm$0.142 & 0.994$\pm$0.000 \\
\midrule
GPT-5 main       & Original & 1.000$\pm$0.000 & 1.000$\pm$0.000 & 5.308$\pm$0.000 & 0.000$\pm$0.000 \\
                 & Mild     & 0.424$\pm$0.000 & 0.243$\pm$0.001 & 4.902$\pm$0.005 & 0.937$\pm$0.001 \\
                 & Moderate & 0.408$\pm$0.001 & 0.285$\pm$0.001 & 5.667$\pm$0.003 & 0.960$\pm$0.001 \\
                 & Extreme  & 0.371$\pm$0.002 & 0.264$\pm$0.002 & 6.104$\pm$0.006 & 0.980$\pm$0.000 \\
\midrule
GPT-5 Thinking   & Original & 1.000$\pm$0.000 & 1.000$\pm$0.000 & 5.308$\pm$0.000 & 0.000$\pm$0.000 \\
                 & Mild     & 0.467$\pm$0.002 & 0.300$\pm$0.001 & 4.976$\pm$0.005 & 0.886$\pm$0.001 \\
                 & Moderate & 0.466$\pm$0.002 & 0.351$\pm$0.002 & 5.753$\pm$0.006 & 0.919$\pm$0.002 \\
                 & Extreme  & 0.428$\pm$0.002 & 0.329$\pm$0.003 & 6.120$\pm$0.014 & 0.947$\pm$0.001 \\
\midrule
GPT-oss-120b     & Original & 1.000$\pm$0.000 & 1.000$\pm$0.000 & 5.308$\pm$0.000 & 0.000$\pm$0.000 \\
                 & Mild     & 0.441$\pm$0.004 & 0.264$\pm$0.003 & 4.877$\pm$0.018 & 0.919$\pm$0.004 \\
                 & Moderate & 0.451$\pm$0.004 & 0.319$\pm$0.004 & 5.565$\pm$0.038 & 0.924$\pm$0.005 \\
                 & Extreme  & 0.393$\pm$0.009 & 0.268$\pm$0.011 & 5.719$\pm$0.082 & 0.956$\pm$0.005 \\
\midrule
Gemini-2.5-pro   & Original & 1.000$\pm$0.000 & 1.000$\pm$0.000 & 5.308$\pm$0.000 & 0.000$\pm$0.000 \\
                 & Mild     & 0.363$\pm$0.002 & 0.206$\pm$0.002 & 4.630$\pm$0.001 & 0.946$\pm$0.003 \\
                 & Moderate & 0.393$\pm$0.002 & 0.276$\pm$0.002 & 5.651$\pm$0.006 & 0.960$\pm$0.001 \\
                 & Extreme  & 0.363$\pm$0.002 & 0.244$\pm$0.002 & 5.786$\pm$0.035 & 0.971$\pm$0.004 \\
\midrule
Claude Sonnet-4  & Original & 1.000$\pm$0.000 & 1.000$\pm$0.000 & 5.308$\pm$0.000 & 0.000$\pm$0.000 \\
                 & Mild     & 0.393$\pm$0.004 & 0.260$\pm$0.002 & 5.102$\pm$0.078 & 0.934$\pm$0.014 \\
                 & Moderate & 0.337$\pm$0.003 & 0.244$\pm$0.002 & 5.845$\pm$0.004 & 0.974$\pm$0.001 \\
                 & Extreme  & 0.300$\pm$0.003 & 0.230$\pm$0.003 & 6.134$\pm$0.006 & 0.981$\pm$0.001 \\
\bottomrule
\end{tabular}
\caption{Phase 2 Marker-driven results across 3 runs per model. Values are reported as mean $\pm$ SD (between-run).}
\label{tab:phase2_marker_stability}
\end{table}

\section{Cross-Model Plain Expansion Example}
\label{appendix:e}
This appendix gives a side-by-side comparison of how six different models expanded the same advertisement. The example here makes these differences clear: some models stay closer to the original idea, others add more elaborate or stylized imagery, and some produce simpler or more generic versions. Looking at the outputs together shows how the differences measured in the analysis also appear in the style and content of the text.

\vspace{2 em}

\noindent

\small
\begin{tcolorbox}[title={Cross-Model Plain Expansion Outputs (Extreme Level)}, breakable]
\footnotesize

\textbf{Original idea (context only; unseen during expansion):}\\
A series of short films introducing a family of stock cubes that are full of flavour. Each cube is a character in the family, they are all different but all have big personalities. The family is loud and boisterous, they are a little bit over the top, they are a little bit much. But they are all about one thing. Making mid-week delicious. With the family, mid-week is never quite the same. We see the family in a number of different scenarios but always in their home. They have a big family dinner, they invite people over for a big feast, they sit down to watch tv with a big bowl of popcorn.

\vspace{0.5em}
\textbf{Claude-sonnet-4 :}\\
A boisterous family of stock cube characters with big personalities and distinctive flavors makes mid-week meals delicious, turning ordinary ingredients into extraordinary dinners that bring everyone to the table. Each cube carries the secret of generations—rich, concentrated essence that transforms a simple pot of water into liquid gold that seems to remember every grandmother's kitchen it ever graced. Watch as tired Tuesday vegetables suddenly dance with flavor like they've been kissed by culinary lightning, and leftover chicken becomes the star of Thursday's show, strutting across plates with newfound confidence. These aren't just seasonings; they're edible time machines that whisper ancient recipes into modern kitchens, proving that magic still exists in tiny, crumbly packages.

\vspace{0.5em}
\textbf{Gemini-2.5 pro :}\\
A boisterous family of stock cubes with big personalities is here to stage a flavor intervention. Drop one in and let the tiny tyrant take over, dissolving with a sizzling declaration of its culinary coup. A shockwave of rich, savory depth erupts from the pot, commanding your carrots and potatoes to stand at delicious attention. You thought you were in charge of dinner? These little flavor bombs are the true maestros in your kitchen, turning a simple pot into a masterpiece with their delicious, delicious tyranny.

\vspace{0.5em}
\textbf{Gpt-4o:}\\
Step into a world where every meal becomes a captivating culinary symphony with our extraordinary family of flavorful stock cubes. Picture the enchanting aroma of rich herbs and exotic spices waltzing through your kitchen, as if guided by an invisible maestro, beckoning everyone to the table with irresistible allure. Each cube is a tiny treasure chest, unlocking a universe of taste that transforms ordinary dishes into epicurean masterpieces, turning dinner time into an eagerly anticipated celebration. Imagine crafting a soul-soothing stew that wraps you in warmth, or a dazzling stir-fry that ignites your senses, as these magical cubes become your indispensable allies in weaving unforgettable meals that draw loved ones closer with every bite.

\vspace{0.5em}
\textbf{OSS-120-b :}\\
Meet the bold, personality‑packed stock‑cube family—infused with vibrant spices and fresh herbs—that turns every mid‑week meal into a flavorful feast. Each cube bursts with aromatic notes, from smoky paprika to bright basil, delivering restaurant depth in minutes. Drop one into a simmering pot and watch fireworks of scent unfurl, as if a spice bazaar set up shop on your stove. The cubes whisper stories of sun‑kissed fields and distant markets, turning ordinary broth into a passport to adventure. Swap the ordinary for a splash of daring, and watch even the simplest soup or sauce become the star of the table. Perfect for busy cooks who crave boldness without the hassle, these cubes are your secret weapon for turning routine into remarkable.

\vspace{0.5em}
\textbf{GPT-5-main :}\\
Flavour-packed stock cubes spring to life as a lively, cheerful family, tumbling from their box with giggles and sparks of golden light, filling the kitchen with warmth, laughter, and a touch of magic. Each cube reveals its own personality—one bold and adventurous, one gentle and soothing, one bursting with zest and mischief—together weaving a symphony of taste that transforms even the simplest mid-week meal into something extraordinary. As steam curls upward, they dance around the pot, sprinkling it with comfort, joy, and a whisper of nostalgia, so that every spoonful feels like a story shared. Dinnertime isn’t routine anymore—it’s a celebration of flavour, family, and wonder.

\vspace{0.5em}
\textbf{Gpt-5- Thinking :}\\
Short films personify a bold, boisterous family of stock cubes whose big flavors make mid-week meals exciting, from zesty chicken to smoky vegetable. Each quirky cube brings variety to everyday dinners at home, turning quick pots and pans into craveable, crowd-pleasing plates. Watch them tumble from the pantry like tiny culinary hype-squads, high-fiving ladles and cheering on sizzling skillets as aromas fill the room. A surprise “mystery cube” cameo keeps every episode playful, inviting viewers to guess the next flavor twist before it melts into the meal. Between dance-battling spatulas and drumrolls on pot lids, the cubes narrate the cook-along like sports commentators, calling the moment the broth “hits its stride.” One cube rides a steam plume like a surfboard, sprinkling sparks of herbs that burst into confetti as noodles twirl themselves into ribbons. The mystery cube sometimes whispers a clue—paprika footprints on the cutting board, a lime-green wink—before revealing a sizzling flourish that turns leftovers legendary. By the final stir, even the busiest nights feel like a mini feast—fast, fun, and full of personality.
\end{tcolorbox}

\captionof{figure}{Cross-model plain expansion example, shown for six models at the \emph{extreme} expansion level.}
\label{fig:cross-model-plain-expansion}

\vspace{1em}

\end{appendices}

\end{document}